\documentclass[11pt]{article}
\usepackage[preprint]{acl}
\usepackage{algorithm}
\usepackage{amsmath} 
\usepackage[most]{tcolorbox}  
\usepackage{fvextra}
\usepackage{fancyvrb} 
\usepackage{adjustbox}
\usepackage{times}
\usepackage{latexsym}
\usepackage[T1]{fontenc}
\usepackage[utf8]{inputenc}
\usepackage{microtype}
\usepackage{inconsolata}
\usepackage{graphicx}
\usepackage{amssymb}
\usepackage{mathtools}
\usepackage{amsthm}
\usepackage{multirow}
\usepackage{booktabs}
\usepackage{tabularx}
\usepackage[capitalize,noabbrev]{cleveref}
\usepackage{xfrac}  
\usepackage{enumitem}
\usepackage{minitoc} %

\noptcrule

\usepackage{array}

\usepackage{algorithmicx}
\usepackage{algpseudocode}

\definecolor{redfg}{HTML}{D35B27}
\definecolor{redbg}{HTML}{FFF7F3}
\definecolor{bluefg}{HTML}{0C69DA}
\definecolor{bluebg}{HTML}{FBFCFE}
\definecolor{Purplefg}{HTML}{7D26CD}
\definecolor{Purplebg}{HTML}{E6E6FA}
\definecolor{greenfg}{HTML}{228B22}
\definecolor{greenbg}{HTML}{F0FFF0}

\definecolor{comm}{gray}{0.5}
\title{\ours: Graph-based Adaptive Tool Evolution Across Diverse Tasks}

\author{
Jianwen Luo\textsuperscript{\rm 1,2}\footnotemark[1],
Yiming Huang\textsuperscript{\rm 1,3}\thanks{Equal contribution.},
Jinxiang Meng\textsuperscript{\rm 1,4,5,6},
Fangyu Lei\textsuperscript{\rm 1,2},\\
\textbf{
Shizhu He\textsuperscript{\rm 1,2},
Xiao Liu\textsuperscript{\rm 3},
Shanshan Jiang\textsuperscript{\rm 7},
Bin Dong\textsuperscript{\rm 7},
Jun Zhao\textsuperscript{\rm 1,2},
Kang Liu\textsuperscript{\rm 1,2}\thanks{Corresponding authors.}}
\\
\textsuperscript{\rm 1}
The Key Laboratory of Cognition and Decision Intelligence for Complex Systems,\\
Institute of Automation, Chinese Academy of Sciences,\\
\textsuperscript{\rm 2}
School of Artificial Intelligence, University of Chinese Academy of Sciences,\\
\textsuperscript{\rm 3}
Microsoft Research Asia,
\textsuperscript{\rm 4}
Nanjing Artificial Intelligence Research of IA,\\
\textsuperscript{\rm 5}
Nanjing University of Posts and Telecommunications,\\
\textsuperscript{\rm 6}
University of Chinese Academy of Sciences, Nanjing,
\textsuperscript{\rm 7}
Ricoh Software Research Center (Beijing)\\
\\
}

\newcommand{\ours}{\scalebox{1.1}{G}ATE}

\begin{document}
\doparttoc
\faketableofcontents

\maketitle

\begin{abstract}
Large Language Models (LLMs) have shown great promise in tool-making, yet existing frameworks often struggle to efficiently construct reliable toolsets and are limited to single-task settings. To address these challenges, we propose \ours\ (\underline{\textit{G}}raph-based \underline{\textit{A}}daptive \underline{\textit{T}}ool \underline{\textit{E}}volution), an adaptive framework that dynamically constructs and evolves a hierarchical graph of reusable tools across multiple scenarios. We evaluate \ours\ on open-ended tasks (Minecraft), agent-based tasks (TextCraft, DABench), and code generation tasks (MATH, Date, TabMWP). Our results show that \ours\ achieves up to 4.3× faster milestone completion in Minecraft compared to the previous SOTA, and provides an average improvement of 9.23\% over existing tool-making methods in code generation tasks and 10.03\% in agent tasks. \ours\ demonstrates the power of adaptive evolution, balancing tool quantity, complexity, and functionality while maintaining high efficiency. Code and data are available at \url{https://github.com/ayanami2003/GATE}.
\end{abstract}

\section{Introduction}
\label{sec:intro}

Large Language Models (LLMs) have demonstrated impressive capabilities in code generation~\citep{cassano2023type, li2023starcoder, roziere2023code, hui2024qwen2}, enabling complex tasks such as mathematical computation~\citep{zhou2023solving, wang2023mathcoder}, tabular reasoning~\citep{chen2022program}, and visual understanding~\citep{suris2023vipergpt, choudhury2023zero}. By generating executable code, LLMs extend their functionality beyond pre-trained parameters, empowering agent-based tasks through frameworks like AutoGen~\citep{wu2023autogen} and CodeActAgent~\citep{wang2024executable}. However, these approaches treat each program as isolated, limiting the reuse of previously generated functional modules across different tasks.

To overcome this, recent studies~\citep{wang2023voyager, cai2023large, qian2023creator, yuan2023craft, stengel2024regal} have focused on developing reusable tool libraries derived from tasks. Despite these advancements, existing methods face significant challenges: (1) \textbf{Toolset Redundancy and Inefficiency}: Many methods generate redundant tools, resulting in bloated libraries that hinder reuse. For example, Voyager~\citep{wang2023voyager} lacks a deduplication mechanism, while CREATOR~\citep{qian2023creator} and CRAFT~\citep{yuan2023craft} create one tool per task, leading to large, repetitive libraries. Regal~\citep{stengel2024regal}, though aiming for simplicity, produces libraries limited to basic arithmetic wrappers. (2) \textbf{Limited Generalizability}: Most methods are validated in narrow settings, restricting their broader applicability. For instance, Voyager~\citep{wang2023voyager} is confined to Minecraft environments, while others~\citep{cai2023large, qian2023creator, yuan2023craft, wang2024trove, stengel2024regal} focus exclusively on code generation tasks.

\begin{figure}[t]
\begin{center}
\centerline{\includegraphics[width=1\linewidth]{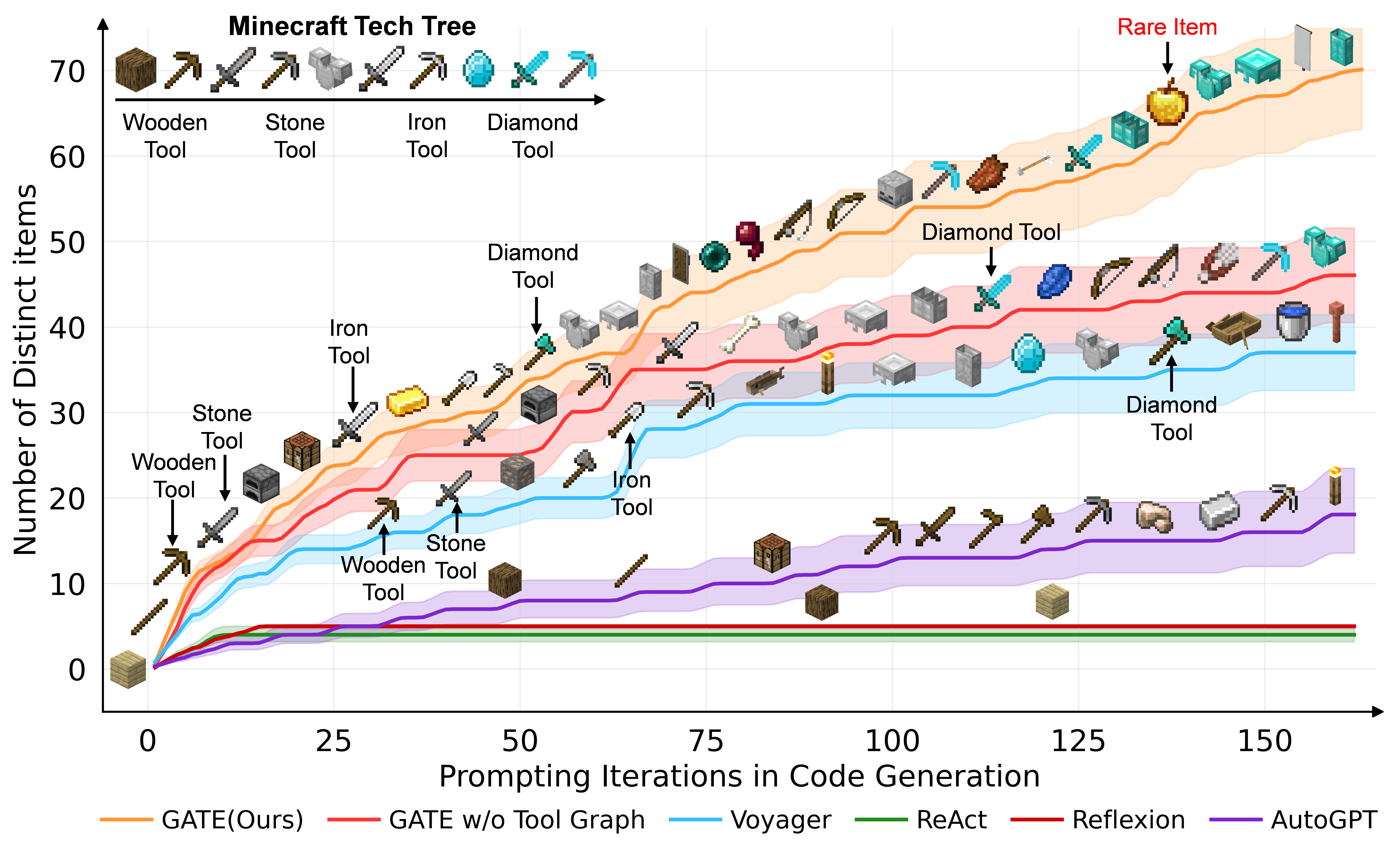}}
\vskip -0.1in
\caption{Performance of \ours\ in Minecraft. \ours\ continually discovers new Minecraft items and skills during exploration, significantly outperforming other methods.}
\label{fig:toolnumber-all}
\end{center}
\vskip -0.3in
\end{figure}
In this paper, we propose \ours\ (\underline{\textit{G}}raph-based \underline{\textit{A}}daptive \underline{\textit{T}}ool \underline{\textit{E}}volution), a framework where two agents, the Task Solver and Tool Manager, dynamically interact with an Adaptive Tool Graph. The Task Solver iteratively extracts tool requirements. The Tool Manager then retrieves tools from the graph using a Graphrank Retrieval method, assembles new tools from existing ones, and refines the graph through pruning and merging. This design sets \ours\ apart from existing tool-making frameworks and addresses the three challenges we discussed as follows: (1) By assembling tools from existing ones instead of generating them from scratch, we improve tool creation efficiency. Additionally, dynamic operations such as merging and pruning ensure the tool library remains concise and manageable. (2) Through the cooperation of two agents, \ours\ dynamically extracts tool requirements based on the current environment and task, converting them into tools. This enables the system to adapt effectively to a wide range of tasks.

We evaluate \ours\ across both open-ended and closed-ended tasks. Our results demonstrate that \ours\ achieves 3.5× better item discovery and 4.3× faster tech tree mastery in Minecraft compared to the previous SOTA (state-of-the-art) method. Additionally, \ours\ outperforms baselines by 5–32\% in agent tasks and surpasses other tool-making methods by an average of 12.6\% in code generation tasks. Our analysis highlights the adaptive evolution of the tool graph across tasks. Compared to other tool-making methods, \ours\ strikes the best trade-off in terms of tool library size, complexity, and redundancy. Our contributions can be summarized as follows:
\begin{itemize}
     \setlength{\itemsep}{0.1em} 
     \item \ours\ is the first method to construct a tool graph by leveraging the invocation relationships between tools, enabling tool evolution and efficient tool retrieval.
     \item \ours\ introduces an agent framework that effectively manages the toolset, maintaining a balanced size with complex tools while avoiding redundancy.
     \item \ours\ achieves generalizability, attaining SOTA performance across various scenarios, including open-ended and closed-ended tasks.
\end{itemize}
   
\begin{figure*}[t]
\begin{center}
\centerline{\includegraphics[width=1\linewidth]{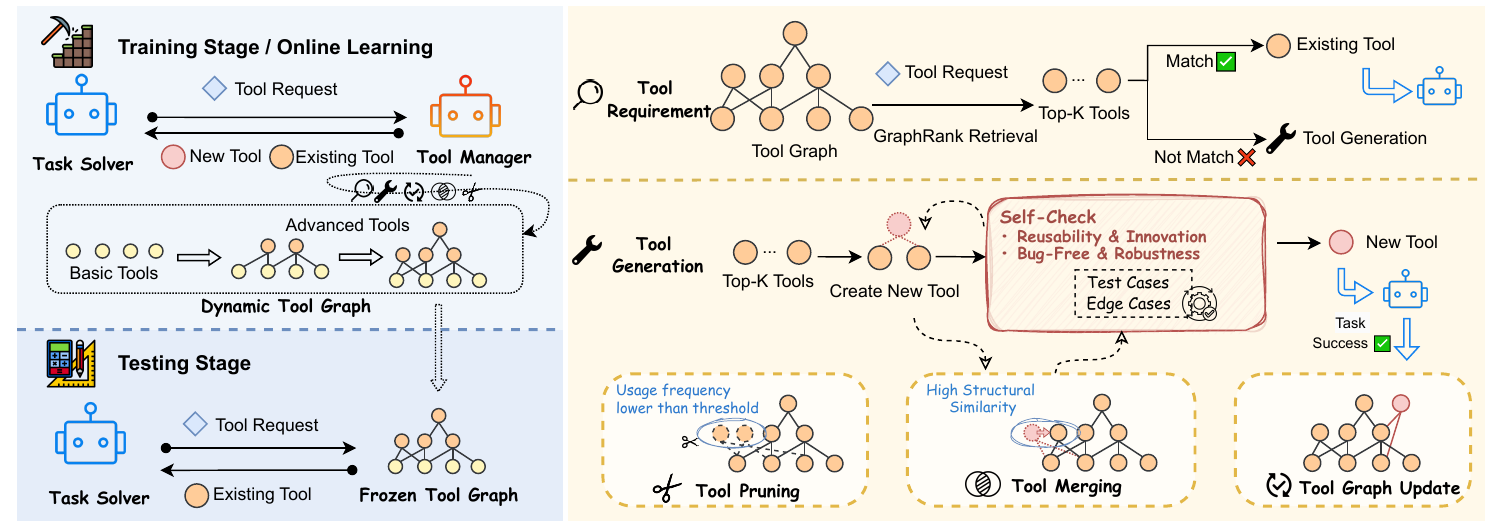}}
\caption{\ours\ consists of two agents: the Task Solver and the Tool Manager, which interact with an Adaptive Tool Graph. Key processes include Tool Requirement, Generation, Pruning, Merging, and Tool Graph Updates.}
\label{fig:toolgraph}
\end{center}
\vskip -0.3in
\end{figure*}

\vspace{-2pt}
\section{Methodology}
\label{sec:method}
\subsection{\ours\ Framework}
As shown in Figure~\ref{fig:toolgraph}, \ours\ consists of two agents: the Task Solver and the Tool Manager, interacting with a dynamic tool graph. Its action space is defined as $\mathcal{A}_s = \{\text{RequestTool}, \text{Terminate}, \text{Code}\}$, allowing it to request tools, operate tools in code, and conclude the task. The Tool Manager assembles or modifies tools based on the Task Solver’s requests, aiming to create high-quality tools. Its action space is given by $\mathcal{A}_t = \{\text{EditTool}, \text{CreateTool}, \text{ReturnTool}\}$, enabling it to edit, create, and return tools. Both agents use the GraphRank algorithm to retrieve tools from the tool graph, with basic tools provided by default.

\subsection{Tool Graph Architecture}
\label{subsec:arch}
\ours 's tool graph is a hierarchical undirected graph, represented as $\mathcal{G} = (\mathcal{V}, \mathcal{E})$, where $\mathcal{V}$ is the set of tool nodes, and $\mathcal{E}$ represents the edges denoting tool dependencies.

\paragraph{Node($\mathcal{V})$} 
The node set $\mathcal{V}$ consists of two types of tools: basic tools, pre-defined by humans, and composed tools, which are created during training. Each node $v_i \in \mathcal{V}$ stores metadata, including the tool's name, docstring, implementation code, usage frequency, and layer position $L(v_i)$. The layer position of basic tools is set to 1, as they form the foundation of the graph. For composed tools, the layer position is determined by the dependencies between the tool and other nodes, where $\text{Call}(v_j, v_i)$ indicates whether tool $v_j$ invokes tool $v_i$ in its implementation:
\begin{equation}
\text{Call}(v_j, v_i) =
\begin{cases}
1, & \text{if } v_j \text{ calls } v_i, \\
0, & \text{otherwise}.
\end{cases}    
\end{equation}
The dependencies of node $v_j$, denoted as $D(v_j) \subset \mathcal{V}$, are given by $D(v_j) = \{ v \in \mathcal{V} \mid \text{Call}(v_j, v) = 1 \}$. The layer position of $v_j$ is then computed as:
\begin{equation}
   L(v_j) = \max\limits_{v \in D(v_j)} L(v) + 1
\end{equation}
\vspace{-3pt}
\paragraph{Edge ($\mathcal{E}$)}
The edge set $\mathcal{E}$ represents the invocation relationships between tools. For any two nodes $v_i, v_j \in \mathcal{V}$, if an invocation relationship exists between them, an edge is established, which is represented through the adjacency matrix $E$. This construction of edges is crucial for capturing the functional dependencies between tools, reflecting how tools interact and depend on each other within the graph. The adjacency matrix $E = \{e_{ij}\}_{N\times N}$ is used to represent these relationships, where $e_{ij}$ is defined as:
\vskip -0.2in
\begin{equation}
    e_{ij} = e_{ji} = \text{Call}(v_i, v_j) \vee \text{Call}(v_j, v_i)
\end{equation}

\subsection{Tool Graph Construction}
As shown in Figure \ref{fig:toolgraph}, \ours involves an iterative collaboration between the Task Solver and the Tool Manager to construct the tool graph. The process begins with the Task Solver extracting \textbf{Tool Requirement}, followed by the Tool Manager entering the Tool Generation phase. This phase consists of three sub-stages: \textbf{Tool Creation}, creating new tools; \textbf{Tool Merging}, which identifies and merges redundant tools; and \textbf{Self-Check}, refining tool effectiveness. The tools are then provided to the Task Solver, and those used in correct solutions are incorporated to \textbf{Update the Tool Graph}.
\vspace{-5pt}
\paragraph{Tool Requirement} Given a task and the current environment, the Task Solver analyzes the situation to extract the required tool requirements and sends them to the Tool Manager. These functionalities are represented as $R = \{r_1, r_2, \dots, r_k\}$. 
For each $r_i$, we use the GraphRank algorithm (see Section \ref{subsec:graphrank}) to retrieve the top-$k$ tools, denoted as $\mathcal{V}_{\text{retrieved}}$, which are then provided to the Tool Manager. 
If any of the $\mathcal{V}_{\text{retrieved}}$ meet the requirement, the Tool Manager directly returns the appropriate tool as $v_i^{\prime}$, minimizing redundant tool creation. Alternatively, if the retrieved tools do not fully satisfy $r_i$, the Tool Manager proceeds to the next stage to create new tools.
\vspace{-5pt}
\paragraph{Tool Creation} The Tool Manager utilizes $\mathcal{V}_{\text{retrieved}}$ to construct new tools, denoted as $\mathcal{V}_{\text{created}}$. Tool creation follows four guiding principles:
 \\
(1) \textbf{Reusability}: Tools should have generalized interfaces and clear names for easy adaptation. \\
(2) \textbf{Leveraging Existing Tools}: Prioritize using retrieved tools for efficiency and modularity. \\
(3) \textbf{Innovation}: New tools should introduce novel functionalities or enhance existing ones. \\
(4) \textbf{Completeness}: Tools must handle edge cases and exceptional inputs to ensure robustness.
\vspace{-5pt}
\paragraph{Tool Merging}
After creating new tools, we assess their potential overlap with existing tools to reduce functional redundancy and enhance the overall tool graph structure. $\mathcal{V}_{\text{created}}$ are compared with the existing tools $\mathcal{V}$ using the Smith-Waterman algorithm~\citep{smith1981identification} to measure structural similarity. The redundant tools of $v_i$ are represented as $\mathcal{R}(v_i)$. If $\mathcal{R}(v_i)$ is not empty, the Tool Manager proceeds to combine the functionalities of $v_i$ and the entire redundant tool set $\mathcal{R}(v_i)$ to generalize a new tool, replacing $v_i$.

\vspace{-5pt}
\paragraph{Self-Check} The Self-Check process evaluates the functionality and quality of created tools in two steps. First, the Tool Manager re-assesses each tool based on the four guiding principles mentioned above. Next, the Tool Manager performs a \textbf{bug-free} verification, generating a few test cases to prevent execution errors. Tools that pass both steps are sent to the Task Solver for integration, while those that fail undergo iterative refinement. The validated tools are represented by $\mathcal{V}_{\text{checked}}$, containing the tools that have passed both checks.
\vspace{-5pt}
\paragraph{Tool Graph Update}
Only correctly solved tasks are considered for updating the tool graph. The tools finally used in these correct solutions are denoted as $\mathcal{V}_{\text{used}} \subset \mathcal{V}_{\text{checked}}$. We then analyze the invocation relationships among the utilized tools, where $v_i \in \mathcal{V}_{\text{used}}$ and $v_j \in \mathcal{V}$, and update the edge set $\mathcal{E}$ and node set $\mathcal{V}$ as follows:
\begin{equation}
{\small
\mathcal{E} \leftarrow \mathcal{E} \cup  
\left\{
(v_i, v_j) \mid \text{Call}(v_i, v_j) = 1
\right\}
}
\end{equation}
\vspace{-2pt}
\begin{equation}
{\small
\mathcal{V} \leftarrow \mathcal{V} \cup \mathcal{V}_{\text{used}}
}
\end{equation}
Finally, we need to remove the corresponding redundant tools $\mathcal{R}(v_i)$ for $v_i \in \mathcal{V}_{\text{used}}$, unless they are used to create a higher-level tool.

\paragraph{Pruning} 
To optimize the tool graph, pruning is performed periodically every $C$ iterations, removing nodes with usage below a threshold $\tau_L$. This threshold is defined as $\tau_l = \lambda \times \log_{10}(C)$. Since higher-level tools tend to be used less frequently, $\lambda$ is adapted based on the tool's level: $\lambda = \frac{1}{1 + 0.8 \times \log_2(L(v_i))}$. To preserve the graph's structural integrity, a rule is enforced: \textit{if a node is non-prunable, all its child nodes are retained}.
\subsection{GraphRank Retrieval}
\label{subsec:graphrank}
To comprehensively capture both semantic similarity and graph structure between tools, we propose \textit{GraphRank Retrieval}, which combines vector similarity retrieval with a modified PageRank algorithm~\citep{xing2004weighted}.
The retrieval process is framed as a random walk on the tool graph, modeled as a Markov chain $\mathcal{M}$ with two key components: the prior probability distribution $p_0$ and the transition matrix $M$. We select the top-k nodes with the highest probabilities from the steady-state distribution $GR$ as the retrieval results.

Given a query and an integer $k$, we first embed the query using \textit{text-embedding-ada-002}~\citep{openai2022textembeddingada002} and compute the cosine similarity $s_i$ between its embedding and each tool’s docstring embedding. These similarity scores are subsequently normalized to get $p_0$:
\begin{equation}
p_0 = \left[ \frac{s_1}{\sum s_i}, \frac{s_2}{\sum s_i}, \dots, \frac{s_N}{\sum s_i} \right] 
\end{equation}
To model the transition probability distribution from each node to others, we treat the distribution as uniform, with the transition probabilities determined by the weight matrix $E$. These probabilities are derived from the column-normalized weight matrix $M=\{m_{ij}\}_{N\times N}$ as follows: 
\begin{equation}
m_{ij} =
\begin{cases}
e_{ij} / \sum_{k=1}^{N} e_{kj} & \text{if } \sum_{k=1}^{N} e_{kj} > 0, \\
1 /N & \text{otherwise}.
\end{cases}
\end{equation}
For isolated nodes, transition probabilities to all nodes are set to $1 / N$, ensuring full participation in the Markov chain. 
Given the probability distribution $GR_{t-1}$ at time step $t-1$, the probability distribution $GR_t$ can be expressed as:
\begin{equation}
    GR_t = (1 - d)p_0 + d \cdot {M}^T GR_{t-1}
\end{equation}
$GR$ satisfies the equation:
\vskip -0.1in
\begin{equation}
GR = (1 - d)p_0 + d \cdot {M}^T GR.
\end{equation}
Here, $d \in [0, 1]$ is a damping factor that balances the influence of the prior distribution and the graph structure. In our implementation, we set $d=0.4$. We directly solve the steady-state equation as a linear system to obtain the solution $GR = (I - d \cdot M^T)^{-1}(1-d)p_0$. The top-$k$ nodes with the highest probabilities in $GR$ are subsequently selected as the retrieved tools.

\begin{table*}[t]
\caption{Mastery of the Tech Tree in the Open-ended Task. The number represents the number of iterations required. Fewer iterations indicate higher efficiency. “N/A” signifies that the number of iterations for obtaining the current tool type is unavailable. Results marked with “*” are from Voyager~\citep{wang2023voyager}.}
\label{tab:Open-Ended Task}
\vskip -0.1in  
\setlength{\tabcolsep}{12pt} 
\renewcommand{\arraystretch}{1} 
\begin{center}
\begin{small}
\begin{sc}
\begin{tabular}{lcccc} 
\toprule
\textnormal{\textbf{Method}} & \textnormal{\textbf{Wood Tool}} & \textnormal{\textbf{Stone Tool}} & \textnormal{\textbf{Iron Tool}} & \textnormal{\textbf{Diamond Tool}} \\
\midrule

\normalfont ReAct$^{*}$ & N/A$(\sfrac{0}{3})$ & N/A $(\sfrac{0}{3})$ & N/A$(\sfrac{0}{3})$ & N/A$(\sfrac{0}{3})$ \\
\normalfont Reflexion$^{*}$ & N/A$(\sfrac{0}{3})$ & N/A$(\sfrac{0}{3})$ & N/A$(\sfrac{0}{3})$ & N/A$(\sfrac{0}{3})$ \\
\normalfont AutoGPT$^{*}$  & 92$\pm$72$(\sfrac{3}{3})$ & 94$\pm$72$(\sfrac{3}{3})$ & 135$\pm$103$(\sfrac{3}{3})$ & N/A$(\sfrac{0}{3})$ \\
\normalfont Voyager & 7$\pm$4$(\sfrac{3}{3})$ & 12$\pm$3$(\sfrac{3}{3})$ & 48$\pm$19$(\sfrac{3}{3})$ & 126$\pm$0$(\sfrac{2}{3})$ \\
\normalfont {\ours}~\textit{\small w/o tool graph}  & 6$\pm$2$(\sfrac{3}{3})$ & 11$\pm$5$(\sfrac{3}{3})$ & 31$\pm$9$(\sfrac{3}{3})$ & 125$\pm$19$(\sfrac{3}{3})$ \\
\normalfont {\ours}~\textit{\small(ours)} & \textbf{4$\pm$0}$(\sfrac{3}{3})$ & \textbf{7$\pm$1}$(\sfrac{3}{3})$ & \textbf{18$\pm$3}$(\sfrac{3}{3})$ & \textbf{29$\pm$2}$(\sfrac{3}{3})$ \\

\bottomrule
\end{tabular}
\end{sc}
\end{small}
\end{center}
\vskip -0.15in
\end{table*}

\begin{table*}[t]
\caption{Test Results of Different Models on the Close-Ended Task. The results are presented for both open-source and closed-source models. “w/o d.” denotes the absence of the tool demo in our method. In the Agent task, “base.” represents ReAct~\citep{yao2022react}, “Refl.” represents Reflexion~\citep{shinn2023reflexion}, and “Plan.” represents Plan-Execution~\citep{shridhar2023art, yang2023intercode}. In the Single-turn Code Tasks, “base.” represents POT~\citep{chen2022program}, and “Crea.” represents CREATOR~\citep{qian2023creator}. $\text{MATH}_{alg}$ represents the algebra subset of MATH, with a difficulty level of 4-5. TabMWP has a difficulty level of 7-8.}
\label{tab:Agent-Ended and signal Task}
\vskip -0.1in
\setlength{\tabcolsep}{3.9pt} 
\renewcommand{\arraystretch}{1} 
\begin{center}
\begin{small}
\begin{sc}
\begin{tabular}{lccccc|lccccccc} 
\toprule

\multicolumn{6}{c|}{\textnormal{\textbf{Multi-turn Agent Tasks}}} & \multicolumn{8}{c}{\textnormal{\textbf{Single-turn Code Tasks}}} \\

\hline

\textnormal{\textit{DS/Mthds}} & \textnormal{\textit{Base.}} & \textnormal{\textit{Refl.}} & \textnormal{\textit{Plan.}} & \textnormal{\textit{Ours}} & \textnormal{\textit{w/o d.}} & \textnormal{\textit{DS/Mthds}} & \textnormal{\textit{Base.}} & \textnormal{\textit{Crea.}} & \textnormal{\textit{Craft}} & \textnormal{\textit{Latm}} & \textnormal{\textit{Regal}} & \textnormal{\textit{Ours}} & \textnormal{\textit{w/o d.}}  \\

\hline

\multicolumn{14}{>{\columncolor{gray!30}}l}{\normalfont \textit{Qwen2.5-7B-Instruct}} \\

\hline
\normalfont TextCraft & 32.90 & 37.60 & 14.53 & \textbf{44.02} & \underline{42.31} & \normalfont $\text{MATH}_{alg}$ & 59.42 & 59.75 & 50.86 & 33.33  & 58.02 & \textbf{73.00} & \underline{69.63} \\ 
\normalfont DA-Bench & 75.77 & \underline{77.78} & 57.99 & \textbf{83.54} & 73.00 & \normalfont Date & 57.59 & 58.33 & 62.45 & 61.57  & 74.81 & \textbf{78.33} & \underline{78.15} \\ 
& & & & & & \normalfont TabMWP
 & 80.57 & 86.38 & 70.32 & 40.72 & 80.91 & \textbf{89.78} & \underline{88.51} \\ 
\hline

\multicolumn{14}{>{\columncolor{gray!30}}l}{\normalfont \textit{Qwen-Coder-7B-Instruct}} \\
\hline
\normalfont TextCraft & 9.40 & 17.52 & 12.82 & \underline{22.22} & \textbf{23.08} & \normalfont $\text{MATH}_{alg}$ & 52.02 & 50.94 & 49.47 & 55.17 & 54.07 & \textbf{69.54} & \underline{63.86} \\
\normalfont DA-Bench & 75.89 & 73.58  & 42.34 & \textbf{81.22} & \underline{76.22} & \normalfont Date  & 61.48 & 57.41 & 61.74 & 52.22 & 74.63 & \underline{78.33} & \textbf{80.55} \\
& & & & & & \normalfont TabMWP & 92.70 & 89.86 & 87.52 & 26.01 & 83.46 & \textbf{95.11} & \underline{93.26} \\ 
\hline

\multicolumn{14}{>{\columncolor{gray!30}}l}{\normalfont \textit{Deepseeker-Coder-6.7B-Instruct}} \\
\hline
\normalfont TextCraft & 2.56 & 11.10 & 6.41 & \textbf{15.38} & \underline{14.10} & \normalfont $\text{MATH}_{alg}$ & 23.95 & 14.34 & 18.52 & 20.38 & 12.10 & \textbf{27.57} & \underline{24.86} \\
\normalfont DA-Bench & 0.63 & 1.27 & 7.59 & \underline{16.78} & \textbf{18.76} & \normalfont Date & 58.89 & 43.88 & 46.85 & 29.61 & 53.89 & \textbf{67.78} & \underline{63.89} \\
& & & & & & \normalfont TabMWP & 70.14 & 81.94 & 66.45  &13.19 & 52.48 & \textbf{87.80} & \underline{82.23} \\ 
\hline

\multicolumn{14}{>{\columncolor{gray!30}}l}{\normalfont \textit{Qwen2.5-14B-Instruct}} \\
\hline
\normalfont TextCraft & 71.79 & 68.37 & 44.87 & \underline{73.93} & \textbf{76.92} & \normalfont $\text{MATH}_{alg}$ & 63.54 & 63.46 & 61.52 & 70.67 & 61.40 & \textbf{77.16} & \underline{74.57} \\
\normalfont DA-Bench & 85.44 & 86.58  & 61.51 & \textbf{87.97} & \underline{86.97} & \normalfont Date & 84.44 & 79.44 & 81.30 & 46.27 & 86.48 & \textbf{88.70} & \underline{87.22} \\
& & & & & & \normalfont TabMWP & 93.19 & 90.49 & 73.29 & 45.58 & 91.13 & \underline{94.68} & \textbf{95.19} \\ 
\hline

\multicolumn{14}{>{\columncolor{gray!30}}l}{\normalfont \textit{Deepseeker-Coder-33B-Instruct}} \\
\hline
\normalfont TextCraft & 8.90 & 12.06 & 2.63 & \textbf{16.67} & \underline{15.28} & \normalfont $\text{MATH}_{alg}$ & 27.45 & 30.62 & 22.13  & \underline{31.90} & 22.13 & \textbf{35.06} & 30.12 \\
\normalfont DA-Bench & 38.46 & 53.79 & 8.22 & \textbf{60.00} & \underline{57.05} & \normalfont Date & 65.00 & 61.85 & 60.00 & 38.43 & 61.11 & \textbf{74.16} & \underline{70.95} \\
& & & & & & \normalfont TabMWP & 83.69 & 89.72 & 80.92 & 22.10 & 80.92 & \textbf{92.76} & \underline{87.45} \\ 
\hline

\multicolumn{14}{>{\columncolor{gray!30}}l}{\normalfont \textit{GPT-3.5-turbo-1106}} \\
\hline
\normalfont TextCraft & 26.92 & 43.59 & 10.27 & \underline{52.85} & \textbf{59.33} & \normalfont $\text{MATH}_{alg}$ & 29.22 & \underline{39.17} & 19.71 & 19.49 & 22.97 & \textbf{42.39} & 34.32 \\
\normalfont DA-Bench & 67.30 & 55.06 & 16.24 & \textbf{72.15} & \underline{71.52} & \normalfont Date  & 71.67 & 66.49 & 61.11 & 55.49 & 73.33 & \textbf{76.85} & \underline{74.44} \\
& & & & & & \normalfont TabMWP & 75.32 & 80.00 & 69.51  & 49.35 & 76.17 & \textbf{83.83} & \underline{82.34} \\
\hline

\multicolumn{14}{>{\columncolor{gray!30}}l}{\normalfont \textit{Claude-3-haiku}} \\
\hline
\normalfont TextCraft & 57.69 & 46.54 & 16.02 & \underline{62.73} & \textbf{66.87}  & \normalfont $\text{MATH}_{alg}$ & 26.34 & 34.16 & \underline{32.59} & 19.05 & 28.48 & \textbf{34.24} & 32.02 \\
\normalfont DA-Bench & 74.68 & 76.16  & 37.39 & \textbf{82.28} & \underline{81.01}  & \normalfont Date & \underline{81.67} & 45.56&74.63&53.33&70.56&\textbf{82.78}& 80.37\\
& & & & & & \normalfont TabMWP & 70.56 & 72.24 & 82.37 & 38.55 & 78.37 & \textbf{90.78} & \underline{90.21} \\
\hline

\multicolumn{14}{>{\columncolor{gray!30}}l}{\normalfont \textit{GPT-4o}} \\
\hline
\normalfont TextCraft & 90.79 & 92.11 & 62.34 & \textbf{96.15} & \underline{94.87}  & \normalfont $\text{MATH}_{alg}$ & 60.98 & 69.13 & 62.22 & 61.94 & 61.73 & \underline{69.80} & \textbf{74.28} \\
\normalfont DA-Bench & 90.16 & 89.69  & 81.43 & \textbf{91.60} & \underline{90.41}  & \normalfont Date & 94.44 & 77.78 &88.33&77.06&93.89&\textbf{95.00}& \underline{95.00}\\
& & & & & & \normalfont TabMWP & 96.60 & 92.98 & 88.96 & 76.74 & \underline{97.66} & \underline{97.66} & \textbf{97.86} \\
\bottomrule
\end{tabular}
\end{sc}
\end{small}
\end{center}
\vskip -0.2in
\end{table*}

\vspace{-2pt}
\section{Experiment Setup}
\label{sec:exp}
We conducted experiments across various scenarios, including both open-ended and closed-ended tasks. We tested traditional single-turn code generation tasks as well as more complex multi-turn agent tasks, covering diverse domains such as games, mathematics, and data science. We briefly introduce the different scenarios in the following section, with further details provided in Appendix \ref{app:apexp}.
\vspace{-5pt}
\subsection{Open-Ended Tasks}
Open-ended tasks~\citep{wang2023voyager} refer to problems that lack a fixed or predefined solution, allowing for multiple possible outcomes. These tasks often require exploration, creativity, and dynamic problem-solving. 
\vspace{-2pt}
\paragraph{Benchmark}
We select Minecraft game as the experimental platform, where players collect resources and craft tools to achieve various objectives. The simulation environment is built on top of Voyager~\citep{wang2023voyager} and leverages Mineflayer~\citep{mineflayer} JavaScript APIs for motor controls. We measure the number of iterations required to complete the tool upgrades, where each code execution for a subtask counts as one iteration.
\vspace{-2pt}
\paragraph{Baselines}
We compare our method with several representative agent algorithms: ReAct~\citep{yao2022react}, Reflexion~\citep{shinn2023reflexion}, AutoGPT~\citep{richardssignificant}, and Voyager~\citep{wang2023voyager}. Some of the experimental results are from Voyager.
\vspace{-5pt}
\paragraph{Implementation}
\ours\ handles open-ended tasks through online learning, where Task Solver continuously addresses ongoing tasks, and Tool Manager adapts the tool graph in real time. \ours\ utilizes GPT-4o for text completion, with tool retrieval limited to 5 and self-checks limited to 2. The bug-free check is omitted to ensure a fair comparison. Tool pruning is performed every 40 steps.

\subsection{Close-Ended Tasks}
Close-ended tasks refer to problems that have a predefined solution or ground truth. We conducted comprehensive experiments, including both single-turn code tasks and multi-turn agent tasks.
\vspace{-5pt}
\paragraph{Benchmark}
For single-turn code tasks, we utilized the algebra subset at levels 4 and 5 from the MATH~\citep{hendrycks2021measuring} dataset, levels 7 and 8 from the TabMWP~\citep{grand2023learning} dataset, and the Date~\citep{srivastava2022beyond} dataset. For multi-turn agent tasks, we performed tests on TextCraft~\citep{cote2019textworld}, a text-based game, and DABench~\citep{hu2024infiagent}, a data science dataset. To prepare the datasets, we selected data for training. The training and testing data amounts are as follows: MATH (200/405), Date (66/180), TabMWP (200/470), TextCraft (98/77), and DABench (98/158). Detailed information on the data splitting methods can be found in Appendix~\ref{app:apexp}. We use the average accuracy for each dataset as the metric. 

\begin{figure}[H]
\vskip -0.05in
\begin{center}
\centerline{\includegraphics[width=0.75\linewidth]{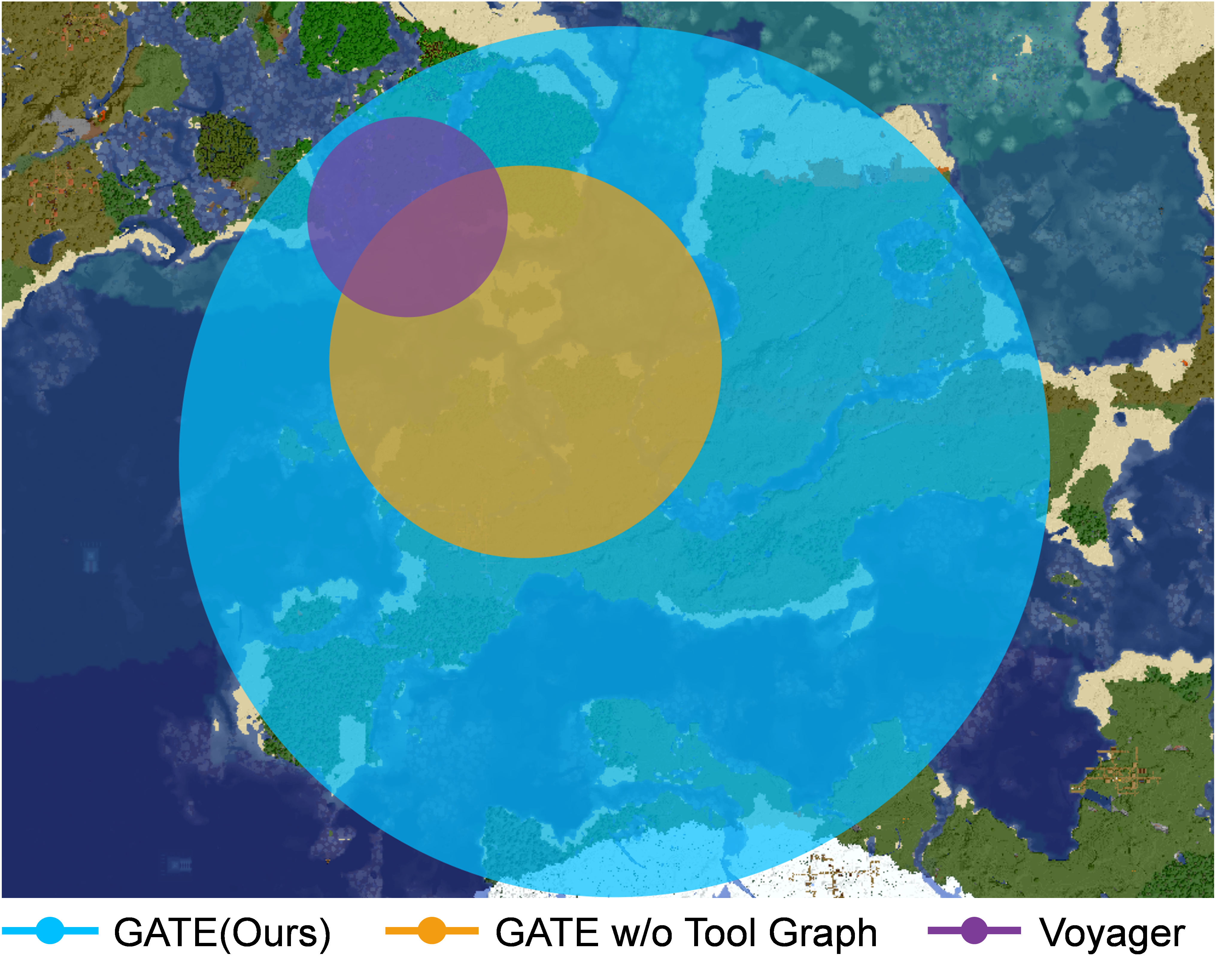}}
\vskip -0.05in
\caption{Map coverage: bird’s eye views of Minecraft maps.}
\label{fig:trial1-map}
\end{center}
\vskip -0.35in
\end{figure}
\vspace{-5pt}
\paragraph{Baselines}
For code tasks, we compare the reasoning framework PoT~\citep{chen2022program} and analyze other tool generation methods, including LATM~\citep{cai2023large}, CREATOR~\citep{qian2023creator}, CRAFT~\citep{yuan2023craft}, and REGAL~\citep{stengel2024regal}. For agent tasks, we compare ReAct~\citep{yao2022react}, Reflexion~\citep{shinn2023reflexion}, and Plan-Execution~\citep{shridhar2023art, yang2023intercode}.
\vspace{-5pt}
\paragraph{Implementation}
For closed-ended tasks, \ours\ separately performs training and testing. During training, \ours\ constructs the tool graph using GPT-4 with greedy decoding, applying tool pruning after training. During testing, the constructed tool graph is frozen, with retrieval enhancing the inference model. Relevant training data and tool code are integrated into the prompt as tool usage examples. For multi-step agent tasks, a ReAct-style~\citep{yao2022react} prompt is employed to facilitate the generation of Thought-Action pairs, whereas single-turn code generation tasks involve direct program synthesis. The complete prompt used is provided in Appendix \ref{appsub:test_prompt}.
\vspace{-5pt}
\paragraph{Models}
For baselines with a tool-making stage, we use GPT-4o as the text completion model. In the test stage, in addition to GPT-4o, we also evaluate several models using constructed tools through the in-context learning method. We test open-source models, including \textit{Qwen2.5-7B-Instruct}~\citep{yang2024qwen2}, \textit{Qwen-Coder-7B-Instruct}~\citep{hui2024qwen2}, \textit{Qwen2.5-14B-Instruct}~\citep{yang2024qwen2}, \textit{Deepseeker-Coder-6.7B-Instruct}~\citep{guo2024deepseek}, and \textit{Deepseeker-Coder-33B-Instruct}~\citep{guo2024deepseek}, while the closed-source models include \textit{GPT-3.5-turbo-1106}, \textit{Claude-3-haiku} and \textit{GPT-4o}. For all experiments, the temperature is set to 0.3, and each experiment is repeated three times, with the average result reported.

\section{Main Results}
\label{sec:mainresult}

\begin{figure*}[ht]
\begin{center}
\centerline{\includegraphics[width=0.88\linewidth]{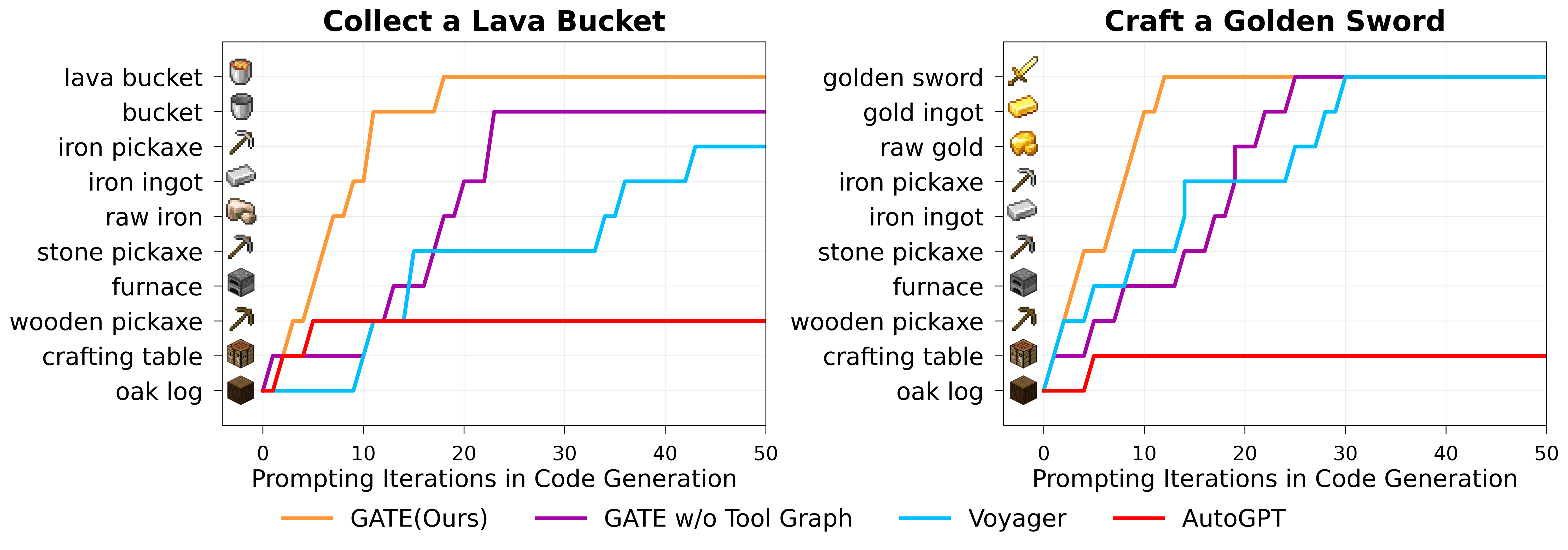}}
\caption{Zero-shot Generalization on Unseen Tasks. The figure visualizes the intermediate progress of each method on two tasks. See Figure \ref{fig:diamon and compass} for the other two tasks. ReAct and Reflexion are excluded from the plot due to their lack of meaningful progress.}
\label{fig:unseen_task}
\end{center}
\vskip -0.3in
\end{figure*}

\paragraph{\ours\ Expands Tech Tree Mastery and Exploration in Open-Ended Tasks.}
\ours\ outperforms the previous SOTA Voyager method in terms of the number of unique items and generates rarer items (Figure \ref{fig:toolnumber-all}). In Minecraft tech tree mastery, \ours\ unlocks the wooden, stone, and iron milestones 23.0×, 13.4×, and 7.5× faster than baselines, respectively (Table \ref{tab:Open-Ended Task}). Notably, \ours\ creates the Diamond Tool 4.34× faster than Voyager and navigates 2.7× longer distances, successfully exploring diverse terrains (Figure \ref{fig:trial1-map}).
\paragraph{\ours\ Enables Self-Improvement on GPT-4o and Boosts Performance on Other Models in Close-Ended Tasks.}
Table \ref{tab:Agent-Ended and signal Task} demonstrates \ours’s effectiveness across both open-source and closed-source models in close-ended tasks. \ours\ facilitates self-improvement on GPT-4o and boosts performance in other models. On average, GPT-4o shows a 5\% improvement in close-ended tasks, while other models achieve gains of 10.03\% and 9.23\% on agent and code sub-tasks, respectively. For instance, GPT-3.5-turbo-1106 improves by 32.4\% on Textcraft, and Qwen2.5-Coder-Instruct sees a 19.07\% increase on Date. These results underscore the adaptability and effectiveness of \ours\ in enhancing performance across various tasks and models.
\paragraph{\ours\ Achieves Significant Improvements Over Other Tool-Making Methods in Close-Ended Tasks.}
As shown in Table \ref{tab:Agent-Ended and signal Task}, \ours\ outperforms other tool-making methods by an average of 10.03\%. Some methodS, such as LATM~\citep{cai2023large} and CRAFT~\citep{yuan2023craft}, perform worse than the baseline model without additional tools, suggesting that their tool libraries may not be as effective. Contrary to the conclusions of CREATOR~\citep{qian2023creator} and CRAFTT~\citep{yuan2023craft}, which separate tool making from tool calling, our results demonstrate that directly generating code yields better performance. 

\begin{figure*}[ht]
\begin{center}
    \begin{minipage}{0.48\textwidth}
        \vspace*{0pt}
        \centering
        \includegraphics[width=\linewidth]{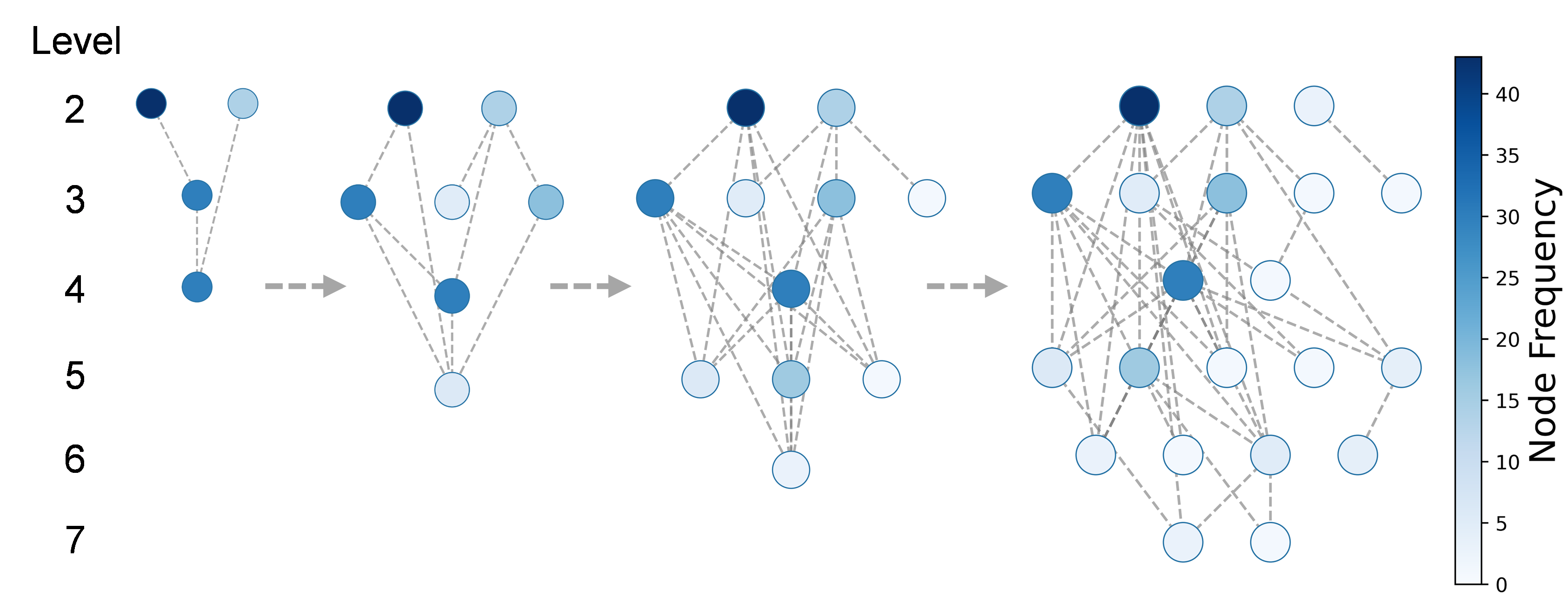}
        \caption{Evolution of the tool graph. We visualize the progression of the tool graph in the Minecraft task, capturing snapshots every 40 steps. The complete evolution for other tasks is provided in the Appendix \ref{subsec:tool-graph}. For clarity, basic tools are excluded from the visualization, as they are generally connected to tools at every level.}
        \label{fig:evlove}
    \end{minipage} 
    \hfill
    \begin{minipage}{0.48\textwidth}
        \vspace*{0pt}
        \centering
        \includegraphics[width=\linewidth]{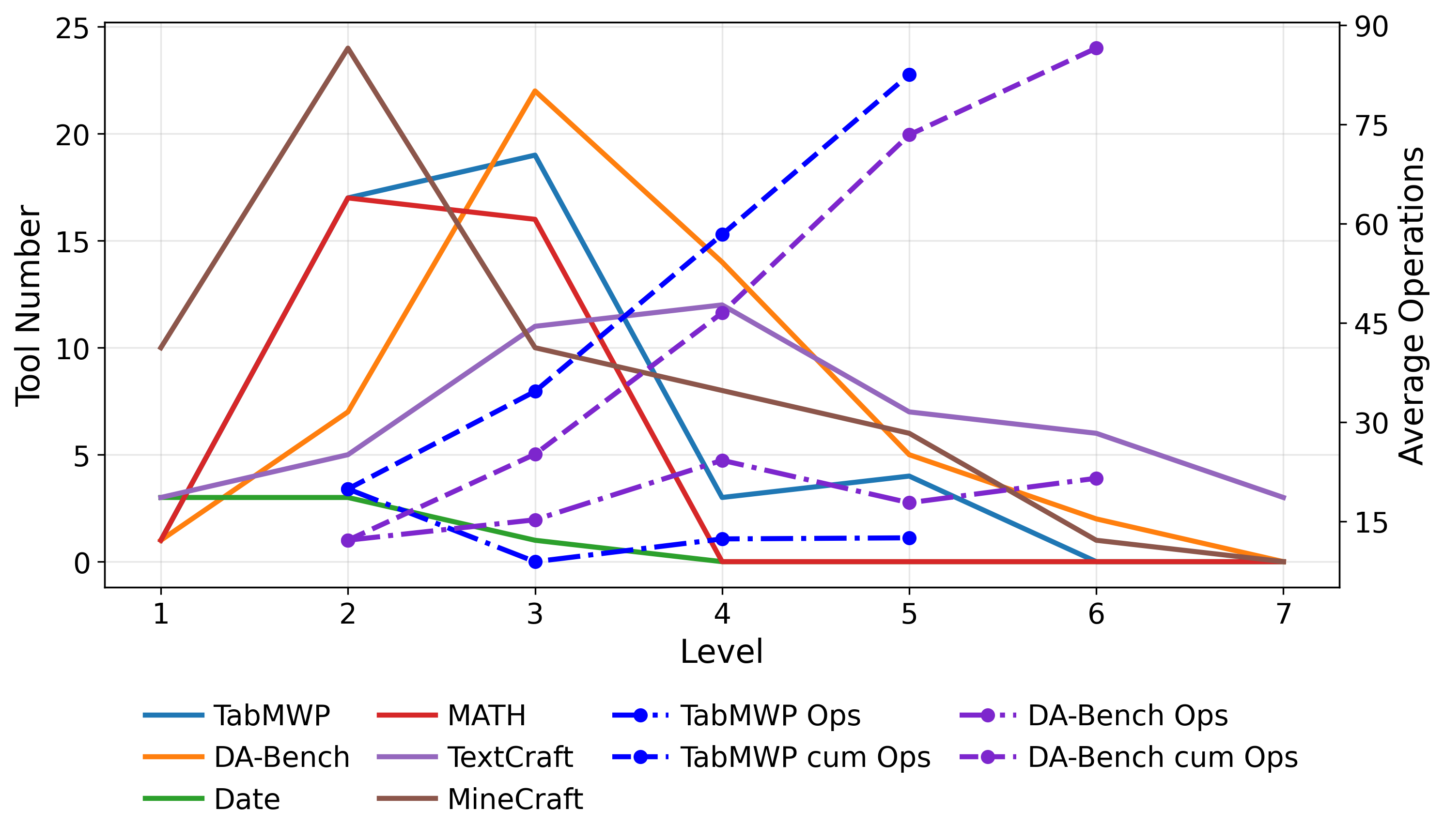}
        \vskip -0.1in
        \caption{Layered Node Distribution of the Tool Graph. "Tool Number" represents the quantity of tools at different levels. The "cum ops" refers to the cumulative number of operations, including function calls.}
        \label{fig:tools}
    \end{minipage}
\end{center}
\vskip -0.3in
\end{figure*}

\section{Analysis}
\label{sec:aly} 
\subsection{How Does \ours\ Adapt to Unseen Tasks?}
To evaluate the generalizability of \ours\ and the effectiveness of the constructed tool graph, we clear the agent's inventory, reset the world to a new instance, and assign previously unseen tasks in Minecraft. The results are summarized in Table \ref{tab:zero_shot} and Figure \ref{fig:unseen_task}.

\begin{table}[H]
\caption{Zero-shot generalization to unseen tasks. We set the maximal prompting iterations as 50. Results with "*" are from Voyager~\citep{wang2023voyager}. "w/o t. g." stands for \textit{without tool graph}.}
\label{tab:zero_shot}
\centering
\resizebox{\columnwidth}{!}{ 
\begin{tabular}{lcccc} 
\toprule
\textbf{Method} & \textbf{Gold Sword} & \textbf{Compass} & \textbf{Diamond Pickaxe} & \textbf{Lava Bucket} \\
\midrule
ReAct$^{*}$ & N/A (\textsuperscript{0}/\textsubscript{3}) & N/A (\textsuperscript{0}/\textsubscript{3}) & N/A (\textsuperscript{0}/\textsubscript{3}) & N/A (\textsuperscript{0}/\textsubscript{3}) \\
Reflexion$^{*}$ & N/A (\textsuperscript{0}/\textsubscript{3}) & N/A (\textsuperscript{0}/\textsubscript{3}) & N/A (\textsuperscript{0}/\textsubscript{3}) & N/A (\textsuperscript{0}/\textsubscript{3}) \\
AutoGPT$^{*}$  & N/A (\textsuperscript{0}/\textsubscript{3}) & N/A (\textsuperscript{0}/\textsubscript{3}) & N/A (\textsuperscript{0}/\textsubscript{3}) & N/A (\textsuperscript{0}/\textsubscript{3}) \\
Voyager & 46$\pm$15 (\textsuperscript{3}/\textsubscript{3}) & 18$\pm$2 (\textsuperscript{3}/\textsubscript{3}) & 22$\pm$4 (\textsuperscript{3}/\textsubscript{3}) & 39 (\textsuperscript{1}/\textsubscript{3}) \\
{\ours}~w/o t. g.  & 33$\pm$20 (\textsuperscript{3}/\textsubscript{3}) & 21$\pm$6 (\textsuperscript{3}/\textsubscript{3}) & 34$\pm$6 (\textsuperscript{3}/\textsubscript{3}) & N/A (\textsuperscript{0}/\textsubscript{3}) \\
{\ours}~(ours) & \textbf{14$\pm$2 (\textsuperscript{3}/\textsubscript{3})} & \textbf{17$\pm$10 (\textsuperscript{3}/\textsubscript{3})} & \textbf{14$\pm$2 (\textsuperscript{3}/\textsubscript{3})} & \textbf{21$\pm$5 (\textsuperscript{3}/\textsubscript{3})} \\
\bottomrule
\end{tabular}
}
\vskip -0.1in
\end{table}

In comparison to Voyager, \ours\ completes tasks 2.2× faster on average. Moreover, when compared to our framework without a tool graph, \ours\ is still 1.8× faster, demonstrating the critical role of the tool graph in enhancing performance. This performance boost highlights the adaptability of \ours\ in handling unseen tasks. By facilitating inter-tool invocation, the tool graph incorporates more comprehensive and generalizable knowledge compared to Voyager’s tool library. This enhanced structure enables \ours\ to generalize across unseen tasks, reinforcing its robustness and versatility in new environments.

\subsection{How Does the Tool Graph Evolve Adaptively?}


As shown in Figure \ref{fig:evlove}, the tool graph evolves dynamically, optimizing its hierarchical structure during training. Initially, it focuses on basic tools (Table \ref{tab:basictool}) and simple relationships, starting with sparse, low-level abstractions. As task complexity increases, tool reuse grows, with frequently used tools becoming key intermediaries.

Figure \ref{fig:tools} illustrates \ours\ adaptive evolution across tasks. For tasks like MATH and TabMWP, which rely on Python libraries, the tool graph remains shallow, with most tools concentrated at lower levels (e.g., 51.5\% of second-level tools in MATH). In contrast, domain-specific tasks like Minecraft and Textcraft lead to deeper, multi-layered graphs, with Textcraft evolving into a 7-layer graph. As the number of layers increases, our higher-level tools save more operations for the same functionality. These patterns highlight the tool graph’s adaptability to task complexity, enabling the extraction of deeper features and the construction of versatile, multi-level tool libraries.

\subsection{How Does the Tool Graph Compare to Other Tool Libraries in Close-Ended TaskS?}
\label{subsec:compare}
Our tool graph framework outperforms existing methods in toolset construction, complexity management, and performance enhancement.  As shown in Table \ref{tab:math_tabmwp_date}, it achieves an optimal balance between tool's complexity (cpl), library size (lib), and average performance improvement (Avg In.) compared to the baselines in Table \ref{tab:Agent-Ended and signal Task}. The tool's complexity (cpl) is calculated by analyzing the Abstract Syntax Tree (AST) of each tool and counting the number of operation nodes, providing a quantitative measure of the tool's complexity. 
\vspace{-1pt}

\begin{table}[h]
    \centering
    \scriptsize
    \setlength{\tabcolsep}{3pt}  
    \caption{Comparison of Tool Libraries Constructed by Different Methods for Single-turn Code Generation Tasks. "Avg In." represents average performance improvement.}
    \label{tab:math_tabmwp_date}
    \begin{tabular}{l p{0.35cm} p{0.3cm} p{0.8cm} p{0.35cm} p{0.3cm} p{0.8cm} p{0.35cm} p{0.3cm} p{0.8cm}}
        \toprule
        \multirow{3}{*}{\textbf{Method}} & \multicolumn{3}{c}{\textbf{$\text{MATH}_{algebra}$}} & \multicolumn{3}{c}{\textbf{TabMWP}} & \multicolumn{3}{c}{\textbf{Date}} \\
        \cmidrule(lr){2-4} \cmidrule(lr){5-7} \cmidrule(lr){8-10}
        & ops & lib & Avg In. & ops & lib & Avg In. & ops & lib & Avg In. \\
        \midrule
        LATM    & -     & -   & -3.9\%       & -     & -   & -43.8\% & -     & -   & -20.2\% \\
        CREATOR & 34.2 & 405 & +2.3\%  & 16.2 & 470 & +2.6\%  & 12.6 & 180 & -10.6\%   \\
        CRAFT   & 9.5  & 138 & -3.2\% & 12.2 & 180 & -5.1\%  & 11.8 & 25  &  -4.9\%  \\
        REGAL   & 4.4 & 8   & -2.8\%     & 5.7  & 7   & -2.7\%   & 4.57  & 9   & +1.69\%       \\
         \textbf{\ours}    & 13.6 & 145 & \textbf{+10.7\%}  & 13.8 & 43  & \textbf{+8.7\%}  & 11.6  & 11   & \textbf{+8.3\%}  \\
        \bottomrule
    \end{tabular}
\end{table}

Compared to CREATOR and CRAFT, \ours\ reduces the number of tools by 86.2\% while improving performance by 9.23\%, demonstrating its ability to construct concise yet highly generalizable tools. REGAL employs pruning to simplify its tool library; however, the resulting toolset has relatively low complexity, with many tools consisting of basic wrappers around library functions or simple foundational operations. Overall, our tool graph offers superior abstraction, generalizability, and efficiency compared to existing methods.

\section{Ablation Studies}
\label{sec:abl}

We conduct ablation studies on four key components—GraphRank Retrieval, Tool Merging, Self-Check, and Pruning—to evaluate their influence on tool graph performance. For Open-ended tasks, we use GPT-4o, and for Close-ended tasks, we train with GPT-4o before testing with \textit{Qwen2.5-14B-Instruct}. Additionally, we use vector-based Top-k retrieval as a baseline to examine the impact of tool graph connectivity. We do not conduct ablation without pruning on open-ended tasks  since \ours\ reaches all milestones before the first pruning, as shown in Table \ref{tab:Agent-Ended and signal Task}. The results are summarized in Table \ref{tab:abl}. 

Among these components, Self-Check and Tool Merging have the greatest impact. Removing Self-Check leads to a 16.3\% accuracy drop in Date and slower tech tree mastery in Minecraft, highlighting its crucial role in validating tool invocation and construction. Tool Merging improves efficiency by reducing redundancy in the tool graph; without it, both task accuracy and the tool graph’s effectiveness suffer. Moreover, GraphRank Retrieval accelerates tool evolution by capturing tool dependencies, demonstrating its importance in streamlining the tool selection process.

\vskip -0.1in
\begin{table}[H]
\caption{Ablation of each optional \ours\ component. \textbf{Above: Open-ended task experiments using GPT-4o.} \textbf{Below: Ablation experiments on the Close-Ended Task}. We selected one dataset from both Agent task and Code generation task, TextCraft and Date. “\textit{W/o Merg.}”,“\textit{W/o Sf.-Chk.}”, “\textit{W/o Prun.}” represent the absence of tool merging, Self-Check, and Pruning, respectively. “\textit{Top-k Ret.}” represents Top-k Retrieval.}
\label{tab:abl}

\centering
\begin{minipage}[t]{0.48\textwidth}
    \centering
    \scriptsize
    \setlength{\tabcolsep}{1.2pt}  
    \renewcommand{\arraystretch}{1} 
    \begin{tabular}{lccccc} 
    \toprule
    \textnormal{\textit{DA/Mthds}}  & \textnormal{\textit{W/o Merg.}} & \textnormal{\textit{W/o Sf.-Chk.}} & \textnormal{\textit{W/o Prun.}} & \textnormal{\textit{Top-k Ret.}} & \textnormal{\textit{{\ours}}} \\
    \midrule
    \normalfont Wood & 6$\pm$2 $(\sfrac{3}{3})$ & 8$\pm$4$(\sfrac{3}{3})$ & - & \textbf{4$\pm$1}$(\sfrac{3}{3})$ & \textbf{4$\pm$0}$(\sfrac{3}{3})$ \\
    \normalfont Stone & 11$\pm$3 $(\sfrac{3}{3})$ & 11$\pm$3$(\sfrac{3}{3})$ & - & \textbf{7$\pm$2}$(\sfrac{3}{3})$ & \textbf{7$\pm$1}$(\sfrac{3}{3})$\\ 
    \normalfont Iron  & 20$\pm$4$(\sfrac{3}{3})$ & 24$\pm$8$(\sfrac{3}{3})$ & - & 25$\pm$7$(\sfrac{3}{3})$ & \textbf{17$\pm$3}$(\sfrac{3}{3})$ \\
    \normalfont Diamond & 56$\pm$9$(\sfrac{3}{3})$ & 52$\pm$0$(\sfrac{1}{3})$ & - & 74$\pm$40$(\sfrac{3}{3})$ & \textbf{29$\pm$2}$(\sfrac{3}{3})$ \\
    \midrule
    \normalfont Textcraft  & 68.09\% & 71.36\% & 69.23\% & 70.56\% & \textbf{73.93\%} \\
    \normalfont Date  & 72.40\% & 71.11\% & 78.89\% & 85.57\% & \textbf{88.70\%} \\
    \bottomrule
    \end{tabular}
\end{minipage}%

\end{table}

\section{Conclusion}
\label{sec:conclusion}
\vspace{-5pt}
In this paper, we introduce \ours, a framework that dynamically constructs a hierarchical tool graph through two-agent collaboration across multiple scenarios. By modeling dependencies between tools and integrating GraphRank for retrieval, \ours\ enables efficient tool discovery, composition, and reuse, effectively addressing key challenges in tool library construction. Experimental results demonstrate that \ours\ outperforms existing methods in both open-ended and closed-ended tasks, achieving superior task-solving accuracy and adaptability. These findings position \ours\ as a robust and scalable solution for autonomous tool-building, paving the way for more advanced agent systems.

\section{Limitations}
\label{sec:lim}
Although our framework has been extended to multiple scenarios, future research should further explore its application to multimodal tasks, such as GUI agents. This would provide a more comprehensive assessment of its generalizability, extending beyond the scope of our current investigation.

While our framework excels in maintaining invocation relationships between tools and in the evolution of basic tools, its ability to construct a complete code project from fundamental components remains to be effectively validated. Future work will be crucial in exploring the boundaries of LLM capabilities and defining the limits of its potential for tool creation.

\bibliography{ref}

\newpage
\appendix
\onecolumn

\part{}
\section*{\centering \LARGE{Appendix}}
\mtcsettitle{parttoc}{Contents}
\parttoc

\clearpage

\section{Related Work}
\label{sec:relatedwork}

\paragraph{Code Generation and Task Solving with LLMs} Large Language Models (LLMs) have demonstrated remarkable potential in generating code to solve complex tasks. Prior studies highlight their effectiveness in mathematical computation ~\citep{zhou2023solving, wang2023mathcoder, gou2023tora}, tabular reasoning ~\citep{chen2022program, lyu2023faithful, lu2024chameleon}, and visual understanding ~\citep{suris2023vipergpt, choudhury2023zero, gupta2023visual}. Frameworks such as AutoGen ~\citep{wu2023autogen} and CodeActAgent~\citep{wang2024executable} extend this capability to agent-based tasks by interpreting executable code as actions. These models dynamically invoke basic tools based on environmental feedback, significantly expanding their utility. Despite their successes, these approaches often treat program generation processes independently, failing to model shared task features and limiting the reusability of functional modules across tasks.

\paragraph{Reusable Tool Creation and Abstraction} To address the limitations of single-use program generation, recent efforts have focused on creating reusable tools. CREATOR ~\citep{qian2023creator} separates the processes of planning (tool creation) and execution, while LATM ~\citep{cai2023large} and CRAFT ~\citep{yuan2023craft} pre-build tools using training and validation sets for task solving. However, these methods often generate a large number of tools, presenting challenges for their efficient reuse. Furthermore, while abstraction-based approaches like REGAL ~\citep{stengel2024regal} focus on extracting reusable tools from primitive programs, they primarily construct simple tools with limited functional complexity. Similarly, Trove ~\citep{wang2024trove} adopts a training-free approach by dynamically composing high-level tools during testing, but its reliance on self-consistency can lead to hallucinated knowledge, reducing accuracy in complex tasks.

\paragraph{Tool Selection for Complex Task Solving} Currently, research on tool selection and retrieval methods primarily focuses on selecting appropriate tools through retrieval mechanisms and LLM-based approaches. ToolRerank ~\citep{zheng2024toolrerank} uses adaptive truncation and hierarchy-aware reranking to improve retrieval results, while Re-Invoke ~\citep{chen2024reinvoketool} introduces an unsupervised framework with synthetic queries and multi-view ranking, enhancing both single-tool and multi-tool retrieval. COLT ~\citep{Qu_2024COLT} combines semantic matching with graph-based collaborative learning to capture relationships among tools, outperforming larger models in some cases. AvaTaR~\citep{wu2024avataroptimizingllmagents} automates the optimization of LLM prompts for better tool utilization, and DRAFT~\citep{qu2024DAFT} refines tool documentation through iterative feedback and exploration, helping LLMs better understand external tools. Despite progress, existing methods generally overlook cost-effectiveness and scalability in tool selection, and often struggle to efficiently adapt to new tools and task requirements in dynamic environments, leading to performance and efficiency bottlenecks. In contrast, our approach dynamically prioritizes tools by combining their relevance and structural importance, ensuring computational efficiency and scalability, thus enabling more effective solutions for complex tasks.
\section{Experimental Details}
\label{app:apexp}
\subsection{Open-ended Task}
\label{subsec:open}
\paragraph{Benchmark} We employed the benchmark proposed by Voyager~\citep{wang2023voyager}, using Minecraft as the experimental platform. Minecraft provides a sandbox environment where players gather resources and craft tools to achieve various goals. The simulation is built on MineDojo~\citep{fan2022minedojo} and uses Mineflayer~\citep{PrismarineJS2013} JavaScript APIs for motor control. 

\paragraph{Baselines}
We conducted a comprehensive comparison with four baselines. Except for Voyager, these methods were originally designed for NLP tasks without embodiment. Therefore, we had to reinterpret and adapt them for execution within the MineDojo environment, ensuring compatibility with the specific requirements of our experimental setup.
\begin{itemize}
    \item \textbf{ReAct:} ReAct~\citep{yao2022react} uses chain-of-thought prompting [46] by generating both reasoning traces and action
plans with LLMs. We provide it with our environment feedback and the agent states as observations.
    \item \textbf{Reflexion:} Reflexion~\citep{shinn2023reflexion} is built on top of ReAct~\citep{yao2022react}with self-reflection to infer more intuitive future actions.
    \item \textbf{AutoGPT:} AutoGPT~\citep{richardssignificant} is a popular software tool that automates NLP tasks by decomposing a high-level
goal into multiple subgoals and executing them in a ReAct-style loop. We re-implement AutoGPT by using GPT-4O to do task decomposition and provide it with the agent states, environment feedback,
and execution errors as observations for subgoal execution
We provide it with execution errors and our self-verification module.
    \item \textbf{Voyager:} Voyager~\citep{wang2023voyager} is a system that integrates an automated curriculum, a scalable skill library, and an iterative prompting framework based on environmental feedback to explore, store, and accumulate skill library within the Minecraft environment.
\end{itemize}

\paragraph{Metric}
The evaluation metric is based on the number of iterations required to progress through tool upgrades, from wooden to stone, iron, and finally diamond tools. Each execution of code is considered one iteration.

\paragraph{Model}
We leverage GPT-4o for text completion, along with the text-embedding-ada-002 API for text embedding. We set all temperatures to
0 except for the automatic curriculum, which uses temperature = 0.1 to encourage task diversity. 

\paragraph{Setting}
We set the maximum number of iterations to 160. For both \ours\ and Voyager, all agents are controlled by GPT-4o, with the number of tools retrieved per iteration set to 5. To ensure a fairer comparison, we removed the Tool Requirement Stage and bug-free checks in \ours\ , and allowed a maximum of 3 self-checks per iteration.

\paragraph{Item Types and Levels}
In the Minecraft task, there are different types and levels of items. Diamond tools are the highest level, and rare items such as golden apples also exist. High-level tools require some lower-level items to craft. Table \ref{tab:toollist} lists the key items in the Minecraft task.
\begingroup
\begin{table}[H]
\caption{List of item types and levels in the Minecraft task.}
\label{tab:toollist}
\vskip -0.1in
\setlength{\tabcolsep}{10pt} 
\begin{center}
\begin{small}
\begin{sc}
\begin{tabular}{l|c|c}
\toprule
\textnormal{\textbf{Category}} & \textnormal{\textbf{level}} & \textnormal{\textbf{Items}} \\
\midrule         
\midrule
\multirow{4}{*}{\multirow{3}{*}{\normalfont Tools}} 
              & \normalfont Wooden Tools & \normalfont Wooden\_Shovel,Wooden\_Pickaxe,Wooden\_Axe,Wooden\_Hoe,Wooden\_Sword \\
              \cmidrule{2-3}
              & \normalfont Stone Tools &\normalfont stone\_pickaxe, stone\_shovel,Stone\_Axe,Stone\_Hoe,Stone\_Sword   \\
              \cmidrule{2-3}
              & \normalfont Iron Tools &\normalfont iron\_pickaxe, iron\_axe, iron\_sword, iron\_shovel, iron\_hoe    \\
              \cmidrule{2-3}
              & \normalfont Diamond Tools &\normalfont diamond\_pickaxe, diamond\_sword, diamond\_axe, diamond\_shovel    \\
             
\midrule
\multirow{2}{*}{\multirow{1}{*}{\normalfont  Armor}} 
              & \normalfont Iron Armor &\normalfont iron\_chestplate, iron\_helmet, iron\_leggings  \\
              \cmidrule{2-3}
              & \normalfont Diamond Armor &\normalfont diamond\_chestplate, diamond\_helmet, diamond\_leggings, diamond\_boots     \\

\midrule
\multirow{3}{*}{\multirow{2}{*}{\normalfont  Food}} 
              & \normalfont Raw Food &\normalfont chicken, mutton, porkchop, rabbit, raw\_rabbit, spider\_eye, bone  \\
              \cmidrule{2-3}
              & \normalfont Cooked Food &\normalfont cooked\_beef, cooked\_chicken, cooked\_mutton, cooked\_porkchop, cooked rabbit  \\
              \cmidrule{2-3}
              & \normalfont Advanced Food &\normalfont golden apple    \\

\bottomrule
\end{tabular}
\end{sc}
\end{small}
\end{center}
\vskip -0.1in
\end{table}
\endgroup

\subsection{Agent Task}
\label{subsec:agent}
\paragraph{Benchmark}
We conducted experiments on two types of agent tasks, demonstrating {\ours}'s capabilities in both game-related and data science tasks.
\begin{itemize}
     \item \textbf{TextCraft:} We evaluate {\ours} on the TextCraft dataset~\citep{futuyma1988evolution}, which challenges agents to craft Minecraft items in a text-only environment~\citep{cote2019textworld}. Each task instance provides a goal and a sequence of crafting commands, which include distractors. We use depth-2 splits for testing and reserve a subset of depth-1 recipes for development, resulting in a 99/77 train/test split.
    \item \textbf{InfiAgent-DABench:} We also test {\ours} on the InfiAgent-DABench benchmark~\citep{hu2024infiagent}, which evaluates LLM-based agents on end-to-end data analysis tasks. This benchmark consists of 257 questions across 52 CSV files, with each question corresponding to a unique CSV file. Agents are required to generate code to analyze data and produce the specified output format. We randomly selected 20 CSV files and their associated question-answer pairs as training data, resulting in a train/test split of 98/159 instances.
\end{itemize}

\paragraph{Baselines}
We compare \ours\ with three methods described below.
\begin{itemize}
     \item \textbf{ReAct:} In this setting, we employ the executor to interact iteratively with the environment, adopting the think-act-observe prompting style from ReAct~\citep{yao2022react}.
     \item \textbf{Plan-Execution:} In contrast, the Plan-and-Execute approach~\citep{shridhar2023art, yang2023intercode} generates a plan upfront and assigns each sub-task to the executor. To ensure each step is executable without further decomposition, we provide new prompts with more detailed planning instructions.
    \item \textbf{Reflexion:} In the Reflection setting~\citep{shinn2023reflexion}, the agent engages in self-reflection after each step, drawing on environmental feedback and exploration history. 
\end{itemize}

\paragraph{Metric} 
The most practically important aspect of the solutions is correctness. For Textcraft, we verify whether the agent’s inventory contains the goal item. For DABench, we check if the agent’s final answer matches the ground truth.

\paragraph{Model}
During training, we use GPT-4o to construct the tool library with a temperature setting of 0. In the testing phase, we conduct a comprehensive comparison of various open-source and closed-source models. The open-source models include \textit{Qwen2.5-7B-Instruct, Qwen-Coder-7B-Instruct, Qwen2.5-14B-Instruct, Deepseeker-Coder-6.7B-Instruct, and Deepseeker-Coder-33B-Instruct}, while the closed-source models primarily include \textit{gpt-3.5-turbo-1106} and \textit{Claude-3-haiku}. During testing, the temperature is set to 0.3, and each experiment is repeated 3 times, with the average result reported.

\paragraph{Setting} 
For ReAct, Reflexion, and \ours\ , the maximum number of steps is set to 20. For Plan-Execution, the maximum number of steps for each sub-task is set to 8. In \ours\ , the number of tools retrieved during testing is limited to 3.

\subsection{Single-turn Code Task}
\label{subsec:code}
\paragraph{Benchmark}
To further explore {\ours}'s potential, we evaluated it on single-turn code generation tasks spanning mathematical reasoning, date comprehension, and tabular reasoning:
 \begin{itemize}
     \item \textbf{MATH:} We used a subset of the MATH dataset~\citep{hendrycks2021measuring}, focusing on 405 level-4 and level-5 algebra problems (MATH contains 5 levels of difficulty) that require textual understanding and advanced reasoning. We randomly selected 200 examples from the test set of the MATH dataset to construct the tool network, resulting in a train/test split of 200/405.
     \item \textbf{Date:} We use the date understanding task from BigBenchHard~\citep{srivastava2022beyond}, which consists of short word problems requiring date understanding. We follow the data splits provided by REGAL\citep{stengel2024regal}, resulting in a train/test split of 66/180.
     \item \textbf{TabMWP:} We further extend our general experiments on MATH by testing on TabMWP~\citep{grand2023learning}, a tabular reasoning dataset consisting of math word problems about tabular data. Based on the CRAFT~\citep{yuan2023craft} splits, we selected 470 problems from levels 7 and 8 (TabMWP contains 8 levels) from the 1,000 test examples. Additionally, we randomly selected 200 examples from the TabMWP training set, resulting in a train/test split of 200/470.
\end{itemize}

\paragraph{Baselines}
For these tasks, we use Programs of Thoughts (PoT)~\citep{chen2022program} and other existing tool-making methods as baselines for comparison.

\begin{itemize}
    \item \textbf{PoT:} The LLM utilizes a program to reason through the problem step by step~\citep{chen2022program}.
   \item \textbf{LATM:} LATM~\citep{cai2023large} samples 3 examples from the training set and create a tool for the task, which is further verified by 3 samples from the validation set. The created tool is then applied to all test cases.
    \item \textbf{CREATOR:} CREATOR~\citep{qian2023creator} disentangle planning (tool making) from execution, enabling Large Language Models (LLMs) to autonomously create a specific tool for each test case during inference.
     \item \textbf{CRAFT:} CRAFT~\citep{yuan2023craft} constructs task-specific toolsets by generating a tool for each training example. During testing, it utilizes a tool retrieval module and a reasoning process akin to CREATOR, generating a function first and then producing the corresponding invocation code. 
      \item \textbf{REGAL:} During training, REGAL~\citep{stengel2024regal} refines primitive programs by extracting functions. In the testing phase, it retrieves both tools and refactored programs—comprising original and refactored versions—to generate a program that effectively solves the task. 
\end{itemize}
\paragraph{Metric}
We use correctness as the evaluation metric, measuring whether the execution outcome of the solution program exactly matches the ground-truth answer(s).
\paragraph{Model}
The models for the single-turn code generation task are the same as those used for the Agent Task, as presented in Section \ref{subsec:agent}.
\paragraph{Setting}
To ensure a fair comparison, we make slight adjustments to each method. For all methods, we allow up to 3 times for format checking and correction, as small models may not always follow the required output format. For PoT, we use 6 fixed examples of basic tool usage as few-shot. CREATOR employs the rectifying process, while for CRAFT, we use the same training set as our method and construct the tool library with GPT-4o, retrieving 3 tools during testing. For Regal, we use PoT along with GPT-4o to obtain ground-truth code, select the correct program, and have GPT-4o reconstruct it. To maintain fairness in tool generation quality, we standardize the few-shot examples of basic tools and retrieve 3 tools, along with 3 usage examples from the current tool library, avoiding errors from pruned tools. For our method, we train with GPT-4o, retrieving 3 tools and their corresponding usage examples during testing, while fixing the basic tool few-shot examples to 3, ensuring consistency with PoT’s total few-shot count.
\section{More Results}
\label{app:apresults}
\subsection{Open-ended Task}
\label{subsec:open-results}
\begin{figure}[H]
\vskip 0.2in
\begin{center}
\centerline{\includegraphics[width=1\linewidth]{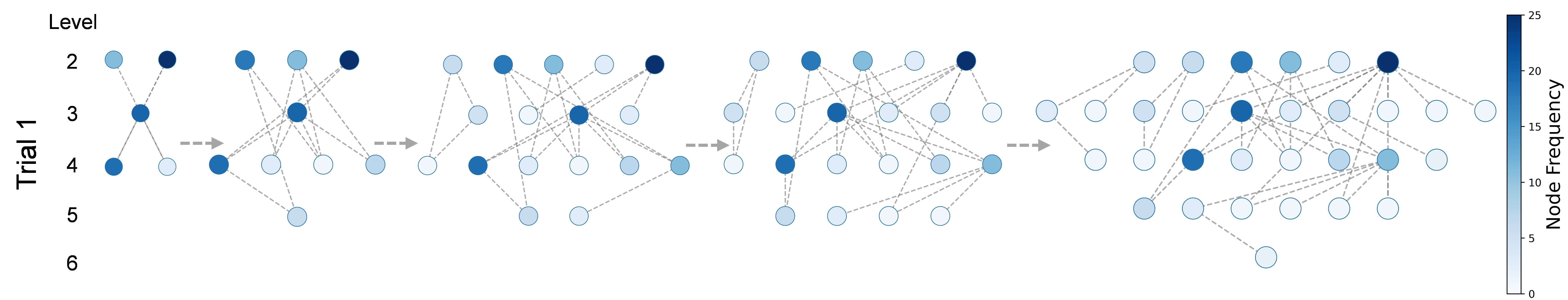}}
\caption{
The tool graph evolution diagram for Minecraft Trial 1. In this diagram, each node represents a tool function, and the edges represent the invocation relationships between tools. The darker the color, the more frequently the tool is invoked. The network consists of a total of 6 layers, with layers 2 to 6 shown here from top to bottom.}
\label{fig:toolnet1}
\end{center}
\vskip -0.3in
\end{figure}

\begin{figure}[H]
\vskip 0.2in
\begin{center}
\centerline{\includegraphics[width=1\linewidth]{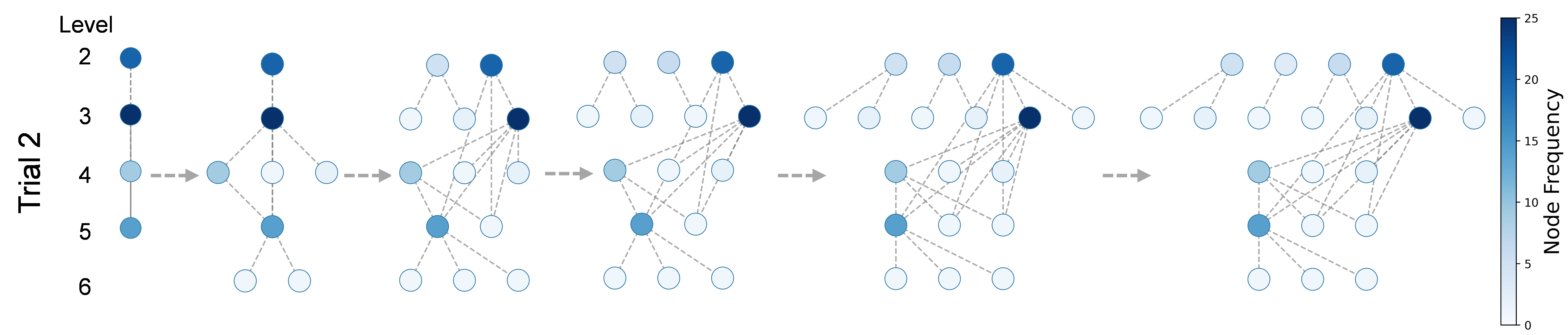}}
\caption{The tool graph evolution diagram for Minecraft Trial 2. In this diagram, each node represents a tool function, and the edges represent the invocation relationships between tools. The darker the color, the more frequently the tool is invoked. The network consists of a total of 6 layers, with layers 2 to 6 shown here from top to bottom.}
\label{fig:toolnet2}
\end{center}
\vskip -0.3in
\end{figure}

\begin{figure}[H]
\vskip 0.2in
\begin{center}
\centerline{\includegraphics[width=1\linewidth]{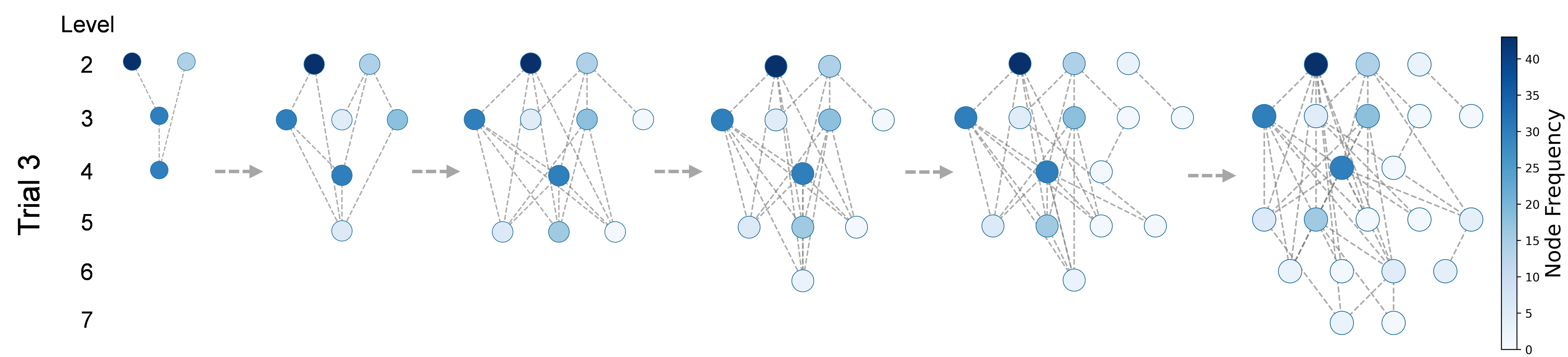}}
\caption{The tool graph evolution diagram for Minecraft Trial 3. In this diagram, each node represents a tool function, and the edges represent the invocation relationships between tools. The darker the color, the more frequently the tool is invoked. The network consists of a total of 6 layers, with layers 2 to 7 shown here from top to bottom.}
\label{fig:toolnet3}
\end{center}
\vskip -0.3in
\end{figure}

\paragraph{More complex tools} 
Our hierarchical graph architecture offers significant advantages in handling complex tasks and large-scale systems. As shown in Figure \ref{fig:toolnet1}, Trial 1 starts with five nodes occupying three layers, and evolves into a five-layer network, with an increasing number of inter-tool calls. As shown in Figure \ref{fig:toolnet2}, Trial 2 starts with four nodes occupying four layers, and evolves into a five-layer network with more inter-tool calls. As shown in Figure \ref{fig:toolnet3}, Trial 3 starts with four nodes occupying three layers, and evolves into a six-layer network structure, with a growing number of inter-tool calls. Our tool graph becomes progressively more complex, flexibly expanding and optimizing its components. These results demonstrate that our method can generate tools that call each other, and combine them into more complex tools. This not only enhances scalability but also facilitates the creation of more sophisticated tools, enabling the solution of increasingly complex problems.

\paragraph{More types of inventory} Our method is able to generate more inventory types than Voyager. As shown in Table \ref{tab:Number}, we can see that {\ours} produces more inventory types in all three trials compared to Voyager.

The inventory collected by {\ours} in each trial is

\begin{itemize}
    \item \textbf{Trial 1:}  \textit{oak\_log, birch\_log, oak\_planks, birch\_planks, crafting\_table, stick, wooden\_pickaxe, dirt, cobblestone, coal, stone\_pickaxe, raw\_copper, furnace, copper\_ingot, andesite, raw\_iron, granite, iron\_ingot, iron\_pickaxe, shield, diorite, raw\_gold, lapis\_lazuli, redstone, diamond, diamond\_pickaxe, bucket, gold\_ingot, iron\_chestplate, arrow, iron\_sword, iron\_helmet, diamond\_sword, diamond\_helmet, lightning\_rod, chest, iron\_axe, iron\_leggings, sandstone, dandelion, spider\_eye, string, iron\_shovel, copper\_block, iron\_door, iron\_hoe, kelp, bow, dried\_kelp, torch, cooked\_beef, gray\_wool, cobbled\_deepslate, tuff, diamond\_leggings, bone, diamond\_chestplate, chicken, white\_banner, cooked\_chicken, egg, feather, oak\_sapling, apple, acacia\_log, golden\_apple, diamond\_axe}

    \item \textbf{Trial 2:}  \textit{oak\_sapling, oak\_log, stick, oak\_planks, crafting\_table, wooden\_pickaxe, dirt, cobblestone, stone\_pickaxe, diorite, raw\_iron, coal, lapis\_lazuli, gravel, furnace, iron\_ingot, raw\_copper, sandstone, granite, iron\_pickaxe, andesite, raw\_gold, gold\_ingot, diamond, diamond\_pickaxe, redstone, cobbled\_deepslate, bucket, iron\_sword, arrow, bow, bone, birch\_log, chest, amethyst\_block, calcite, smooth\_basalt, iron\_chestplate, diamond\_sword, diamond\_helmet, iron\_leggings, diamond\_boots, water\_bucket, string, orange\_tulip, mutton, white\_wool, porkchop, dandelion, cooked\_porkchop, cooked\_mutton}

    \item \textbf{Trial 3:}  \textit{jungle\_log, stick, oak\_sapling, jungle\_planks, crafting\_table, dirt, wooden\_pickaxe, cobblestone, stone\_pickaxe, raw\_iron, raw\_copper, furnace, iron\_ingot, iron\_pickaxe, coal, diorite, lapis\_lazuli, andesite, moss\_block, clay\_ball, redstone, raw\_gold, cobbled\_deepslate, granite, diamond, diamond\_pickaxe, copper\_ingot, gunpowder, bucket, gravel, gold\_ingot, oak\_log, iron\_sword, iron\_chestplate, chest, diamond\_sword, spruce\_sapling, rotten\_flesh, bone, rose\_bush, water\_bucket, string, oak\_planks, grass\_block, diamond\_helmet, iron\_leggings, emerald, snowball, rabbit\_hide, rabbit, spruce\_log, cooked\_rabbit, diamond\_boots}
\end{itemize}

The inventory collected by Voyager in each trial is
\begin{itemize}
    \item \textbf{Trial 1:}  \textit{oak\_log, birch\_log, oak\_sapling, birch\_sapling, oak\_planks, stick, crafting\_table, wooden\_pickaxe, dirt, cobblestone, stone\_pickaxe, raw\_copper, white\_tulip, coal, furnace, copper\_ingot, granite, raw\_iron, iron\_ingot, lightning\_rod, iron\_pickaxe, pink\_tulip, orange\_tulip, sandstone, shears, shield, diorite, cobbled\_deepslate, iron\_block, chest, tuff, lapis\_lazuli, redstone, diamond, raw\_gold, gold\_ingot, diamond\_pickaxe, diamond\_helmet, diamond\_sword, sand, andesite, arrow, bone, iron\_chestplate, beef, leather, oak\_leaves, porkchop, cooked\_beef, leather\_leggings}

    \item \textbf{Trial 2:}  \textit{dirt, oak\_log, oak\_planks, crafting\_table, stick, oak\_sapling, wooden\_pickaxe, cobblestone, coal, stone\_pickaxe, raw\_iron, granite, lapis\_lazuli, raw\_copper, furnace, iron\_ingot, copper\_ingot, iron\_helmet, iron\_pickaxe, diorite, andesite, salmon, ink\_sac, iron\_chestplate, lightning\_rod, cooked\_salmon, stone, stonecutter, rotten\_flesh, gravel, flint, chest, iron\_leggings, copper\_block, cobbled\_deepslate, tuff, diamond, diamond\_pickaxe, raw\_gold, gold\_ingot, redstone, diamond\_sword, egg, diamond\_boots, diamond\_axe}

    \item \textbf{Trial 3:}  \textit{jungle\_log, jungle\_planks, oak\_sapling, oak\_log, crafting\_table, stick, wooden\_pickaxe, dirt, cobblestone, coal, stone\_pickaxe, raw\_copper, furnace, copper\_ingot, magma\_block, lightning\_rod, stone\_axe, jungle\_boat, kelp, sand, sandstone, glass, raw\_iron, granite, lapis\_lazuli, diorite, iron\_ingot, bucket, iron\_pickaxe, chest, andesite, redstone, dried\_kelp, iron\_chestplate, wooden\_sword, shield, iron\_sword}
\end{itemize}

\vskip -0.2in
\begin{table}[H]
\caption{Number of different inventory types produced by each trial}
\label{tab:Number}
\setlength{\tabcolsep}{12pt} 
\renewcommand{\arraystretch}{1.0} 
\begin{center}
\begin{small}
\begin{sc}
\begin{tabular}{lccc} 
\toprule
\textnormal{\textbf{Method}} & \textnormal{\textbf{Trial 1}} & \textnormal{\textbf{Trial 2}} & \textnormal{\textbf{Trial 3}}  \\
\midrule
\normalfont Voyager     & 50  & 45  & 37    \\
\normalfont AETG(Ours)  & 67  & 51  & 53    \\
\bottomrule
\end{tabular}
\end{sc}
\end{small}
\end{center}
\vskip -0.1in
\end{table}

\begin{figure}[H]
\vskip 0.2in
\begin{center}
\centerline{\includegraphics[width=1\linewidth]{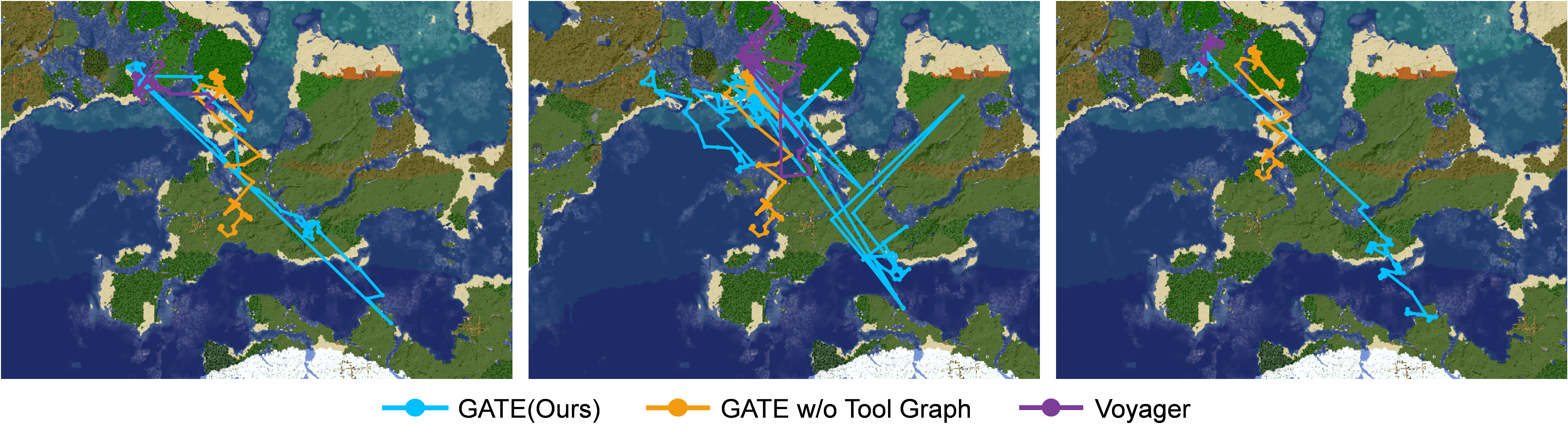}}
\caption{Movement trajectory Map: Three bird’s eye views of Minecraft maps. The trajectories are plotted based on the position coordinates where each agent interacts.}
\label{fig:linermap}
\end{center}
\vskip -0.3in
\end{figure}

\paragraph{Longer exploration path} To better demonstrate the exploration capabilities of the agent, we compared the exploration trajectories and their lengths. As shown in Figure \ref{fig:linermap}, our agent exhibits longer and more persistent exploration capabilities than Voyager. In Table \ref{tab:length}, the trajectory lengths of our agent are consistently much greater than those of Voyager. {\ours}is able to traverse across multiple terrains, with an average distance 2.66 times longer than Voyager. Additionally, {\ours} can explore across different continental plates, while Voyager remains confined to a single plate, highlighting the exceptional exploration capability of {\ours}.

\begin{table}[H]
\caption{Exploration trajectory length in each trial, where \textit{Performance Gain} = $\textit{ours}/\textit{voyager}$.}
\label{tab:length}
\setlength{\tabcolsep}{12pt} 
\begin{center}
\begin{small}
\begin{sc}
\begin{tabular}{lcccc} 
\toprule
\textnormal{\textbf{Method}} & \textnormal{\textbf{Trial 1}} & \textnormal{\textbf{Trial 2}} & \textnormal{\textbf{Trial 3}} & \textnormal{\textbf{\textit{Avg}}}\\
\midrule
\normalfont Voyager     & 1925.74  & 4102.99  & 902.13  & 2310.29   \\
\normalfont {\ours}(Ours)  & 5665.75  & 8908.57  & 3895.06 & 6156.46  \\
\midrule
\normalfont \textit{Performance Gain} & 2.94  & 2.17   & 4.32    & 2.66 \\
\bottomrule
\end{tabular}
\end{sc}
\end{small}
\end{center}
\vskip -0.1in
\end{table}

\vskip -0.2in
\begin{figure}[H]
\vskip 0.2in
\begin{center}
\centerline{\includegraphics[width=1\linewidth]{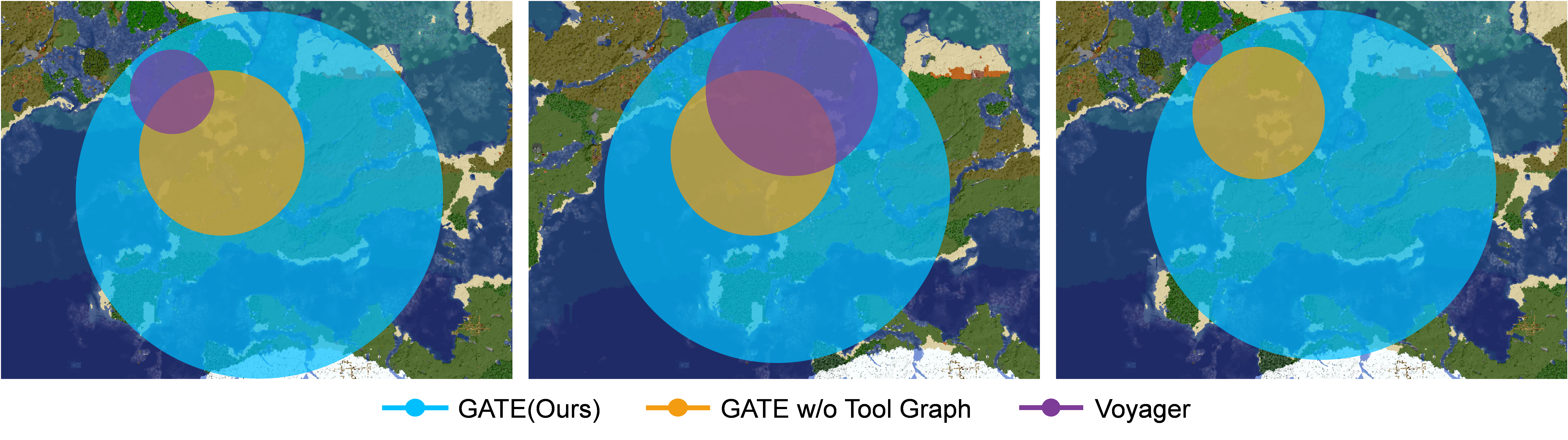}}
\caption{Map coverage: Three bird’s eye views of Minecraft maps. The trajectories are plotted based on the position coordinates where each agent interacts.}
\label{fig:trialmap}
\end{center}
\vskip -0.3in
\end{figure}

\vskip -0.2in

\begin{figure}[H]
\vskip 0.2in
\begin{center}
\centerline{\includegraphics[width=1\linewidth]{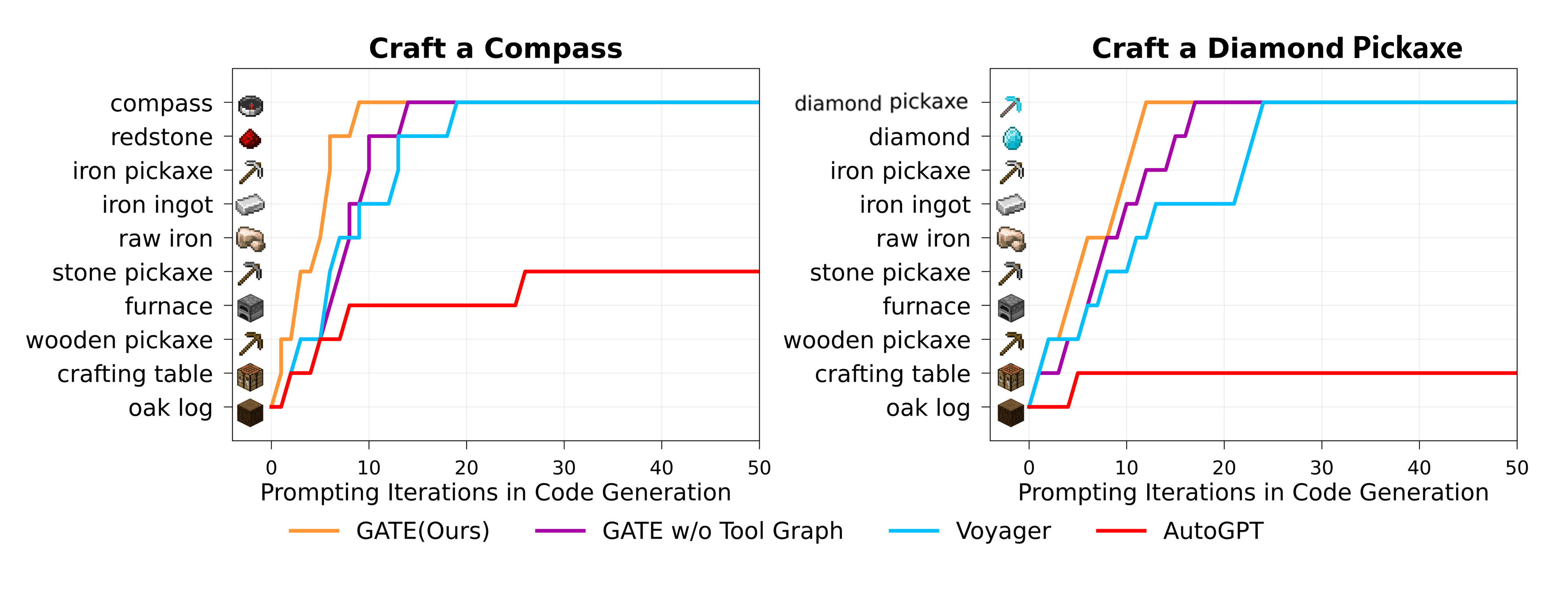}}
\caption{Zero-shot generalization to unseen tasks. Here, we visualize the intermediate progress of each method on the tasks "Craft a Compass" and "Craft a Diamond Hoe."}
\label{fig:diamon and compass}
\end{center}
\vskip -0.3in
\end{figure}

\paragraph{Efficient Zero-Shot Generalization to Unseen Tasks} Based on the results presented in Table \ref{tab:newtechtree} and Figure \ref{fig:diamon and compass}, we can clearly observe the significant advantages of {\ours} in the open-ended task. Table \ref{tab:newtechtree} shows the number of iterations required for different methods to complete various tasks (Gold Sword, Compass, Diamond Hoe, Lava Bucket), where fewer iterations indicate higher efficiency. Compared to Voyager and {\ours} (w/o toolnet), {\ours} consistently requires significantly fewer iterations across all tasks, demonstrating substantial improvements in efficiency. Notably, in the Gold Sword task, {\ours} (ours) completes the task in just 14.00±1.73 iterations, whereas Voyager requires 46.33±14.57 iterations, showcasing its superior performance.

Figure \ref{fig:diamon and compass} further visualizes the intermediate progress of different methods on the "Craft a Compass" and "Craft a Diamond Hoe" tasks. It is evident that {\ours} learns and masters the necessary skills for crafting items more quickly. As the number of prompting iterations increases, {\ours} reaches the task objectives significantly earlier than the other methods. Additionally, while {\ours}(w/o Tool Graph) performs better than Voyager, it still lags behind {\ours}, indicating that the ToolNet component plays a crucial role in enhancing the model's capability.

Overall, these experimental results demonstrate that {\ours} not only learns new skills and crafting techniques more efficiently but also that its key module, Tool Graph, is essential for overall performance improvement. This further validates the effectiveness of our approach in self-driven exploration and task generalization.

\begingroup
\begin{table}[H]
\caption{The mastery of the tech tree in the Open-ended Task. The number indicates the number of iterations. The fewer the iterations, the more efficient the method. "N/A" indicates that the number of iterations for obtaining the current type of tool is not available.}
\label{tab:newtechtree}
\vskip 0.1in
\setlength{\tabcolsep}{12pt} 
\begin{center}
\begin{small}
\begin{sc}
\begin{tabular}{lccccc} 
\toprule
\textnormal{\textbf{Method}} & \textnormal{\textbf{Trial}} & \textnormal{\textbf{Gold Sword}} & \textnormal{\textbf{Compass}} & \textnormal{\textbf{Diamond Pickaxe}} & \textnormal{\textbf{Lava Bucket}} \\
\midrule
\multirow{4}{*}{\multirow{2}{*}{\normalfont Voyager}} 
              & \normalfont Trial 1 & 48 & 16 &  24 & N/A         \\
              & \normalfont Trial 2 & 31 & 17 &  25 & 39         \\
              & \normalfont Trial 3 & 60 & 20 & 18  & N/A         \\
              \cmidrule{2-6}
              & \textit{Average} & 46.33$\pm$14.57 & 17.67$\pm$2.08 & 22.33$\pm$3.79 & 39.00$\pm$0.00 \\
\midrule
\multirow{4}{*}{\multirow{2}{*}{\normalfont {\ours}\textit{\small(w/o toolnet)}}} 
               & \normalfont Trial 1 & 26 & 27 & 23  & N/A         \\
              & \normalfont Trial 2 & 18 & 22 & 18  & N/A        \\
              & \normalfont Trial 3 & 56 & 15 & 30  & N/A          \\
              \cmidrule{2-6}
              & \textit{Average} & 33.33$\pm$20.03 & 21.33$\pm$6.03 & 23.67$\pm$6.03 & N/A$\pm$N/A \\
\midrule
\multirow{4}{*}{\multirow{2}{*}{\normalfont {\ours}\textit{\small(ours)}}} 
              & \normalfont Trial 1 & 13 & 28 & 16  & 19       \\
              & \normalfont Trial 2 & 13 & 10 & 14  & 27       \\
              & \normalfont Trial 3 & 16 & 13  & 13  & 18      \\
              \cmidrule{2-6}
              & \textit{Average} & \textbf{14.00$\pm$1.73} & \textbf{17.00$\pm$9.64} & \textbf{14.33$\pm$1.53} & \textbf{21.33$\pm$4.93} \\

\bottomrule
\end{tabular}
\end{sc}
\end{small}
\end{center}
\vskip -0.1in
\end{table}
\endgroup

\subsection{Agent Task}
\label{subsec:agent-results}

\begin{figure}[H]
\vskip 0.2in
\begin{center}
\centerline{\includegraphics[width=1\linewidth]{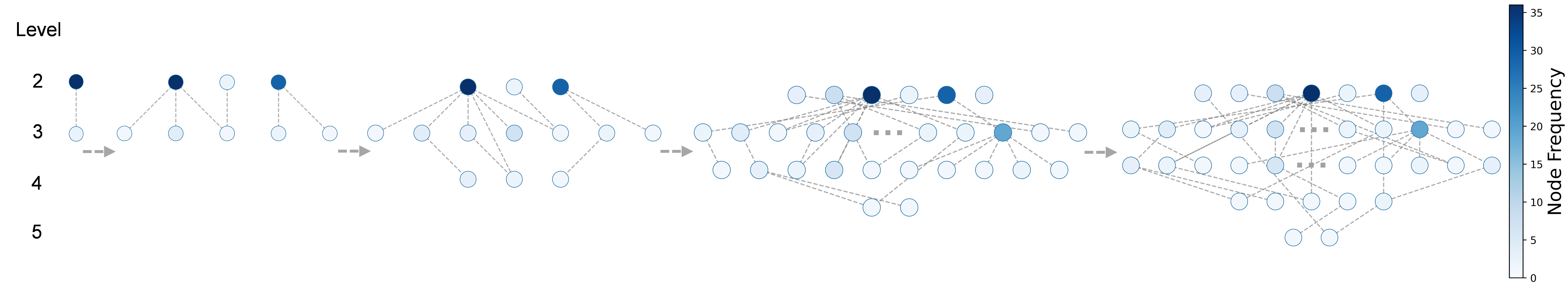}}
\caption{The tool graph evolution diagram of DA-Bench. In this diagram, each node represents a tool function, and the edges represent the invocation relationships between tools. The darker the color, the more frequently the tool is invoked.}
\label{fig:toolnet-dabench}
\end{center}
\vskip -0.3in
\end{figure}

\begin{figure}[H]
\vskip 0.2in
\begin{center}
\centerline{\includegraphics[width=1\linewidth]{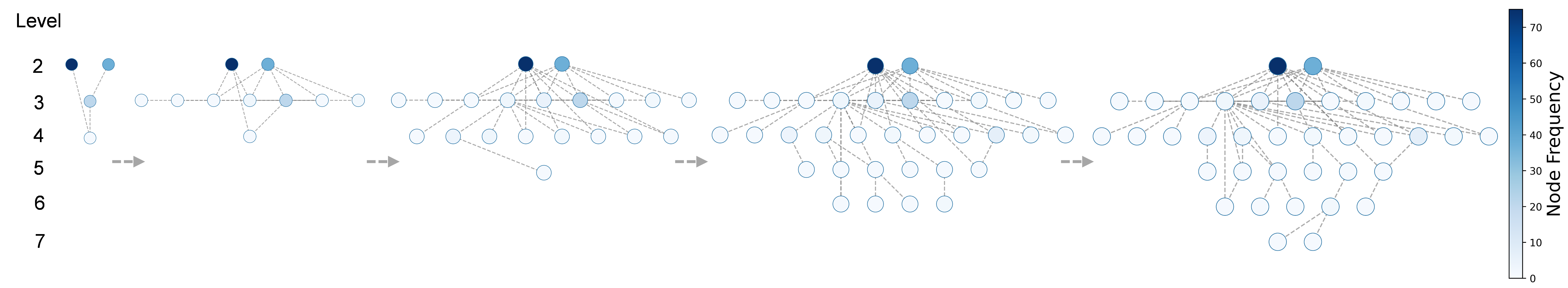}}
\caption{The tool graph evolution diagram of TextCraft. In this diagram, each node represents a tool function, and the edges represent the invocation relationships between tools. The darker the color, the more frequently the tool is invoked.}
\label{fig:toolnet-textcraft}
\end{center}
\vskip -0.3in
\end{figure}

Figures \ref{fig:toolnet-dabench} and \ref{fig:toolnet-textcraft} present the tool network evolution diagrams of DA-Bench and TextCraft, which visually reflect the call relationships between different tool functions. In these diagrams, each node represents a specific tool function, edges indicate the call dependencies between tools, and the shading of the nodes reflects the frequency of tool calls—darker colors indicate higher call frequency. From Figure \ref{fig:toolnet-dabench}, it can be observed that in DA-Bench, the tool network expands progressively as the task advances, forming multiple core nodes with higher call frequencies. This suggests that certain key tools are frequently called during the task execution, playing a central role. Additionally, the tool call relationships exhibit a hierarchical and well-organized structure, reflecting DA-Bench's efficiency in tool dependency management.

In contrast, Figure \ref{fig:toolnet-textcraft} illustrates the tool network evolution of TextCraft, which also shows a similar expansion trend overall. However, compared to DA-Bench, the tool call frequency in TextCraft is more evenly distributed across multiple nodes, meaning that the system calls a wider variety of tools during task execution, rather than relying on a few core tools. This distribution pattern may suggest that TextCraft adopts a more diverse tool usage strategy in task execution.

A comparative analysis of the two figures reveals that, although both DA-Bench and TextCraft exhibit certain hierarchical and expansive characteristics in their tool call patterns, DA-Bench relies more heavily on a few core tools, whereas TextCraft displays a more dispersed tool call pattern. This contrast not only highlights the differences in tool usage between the two, but also emphasizes the importance and effectiveness of ToolNet.

\subsection{Single-turn Code Task}
\label{subsec:code-results}

\begin{figure}[H]
\vskip 0.2in
\begin{center}
\centerline{\includegraphics[width=1\linewidth]{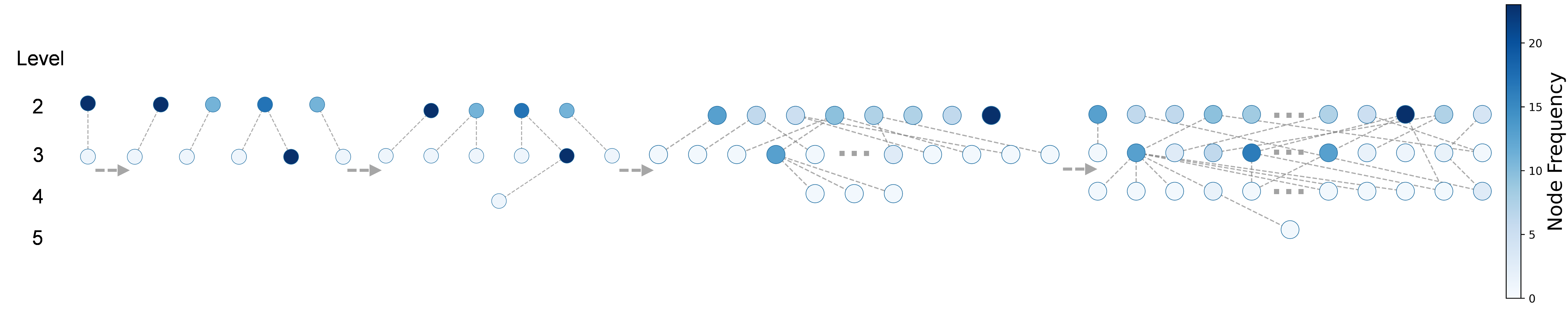}}
\caption{The tool graph evolution diagram of MATH. In this diagram, each node represents a tool function, and the edges represent the invocation relationships between tools. The darker the color, the more frequently the tool is invoked.}
\label{fig:toolnet-math}
\end{center}
\vskip -0.3in
\end{figure}

\begin{figure}[H]
\vskip 0.2in
\begin{center}
\centerline{\includegraphics[width=1\linewidth]{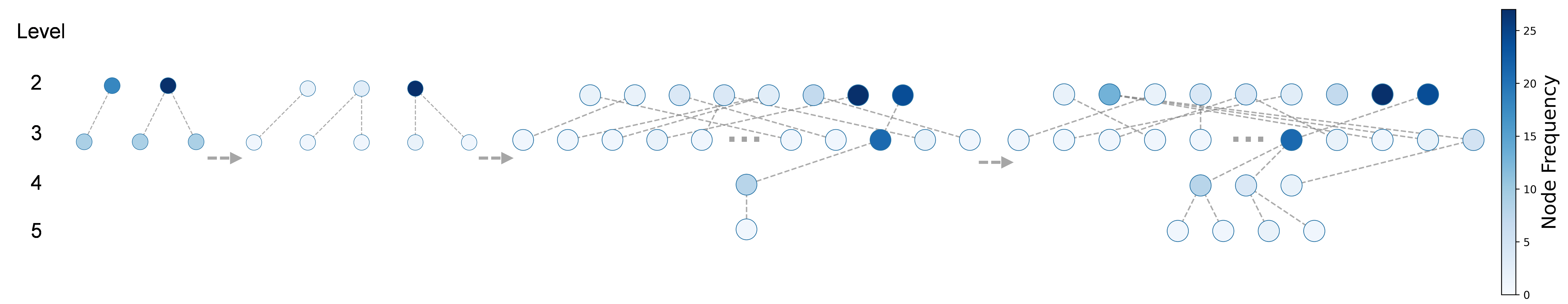}}
\caption{The tool graph evolution diagram of TabMWP. In this diagram, each node represents a tool function, and the edges represent the invocation relationships between tools. The darker the color, the more frequently the tool is invoked.}
\label{fig:toolnet-tabmwp}
\end{center}
\vskip -0.3in
\end{figure}

As shown in the Figure\ref{fig:toolnet-math} \ref{fig:toolnet-tabmwp}, this illustrates the evolution of the tool graph for the Math and TabMWP tasks. It is evident that the tool graph gradually becomes more complex, creating multiple layers of tools, making the tool graph more intricate. Since the Date task can be solved with fewer tools, there is no evolution of the tool graph. However, the generated tools can still effectively solve the task, while there exists a multi-level calling relationship.

\section{More Ablations}
\label{app:apablation}
\subsection{Open-ended Task}
\label{subsec:open-ablation}

\begin{figure}[H]
\begin{center}
\centerline{\includegraphics[width=0.6\linewidth]{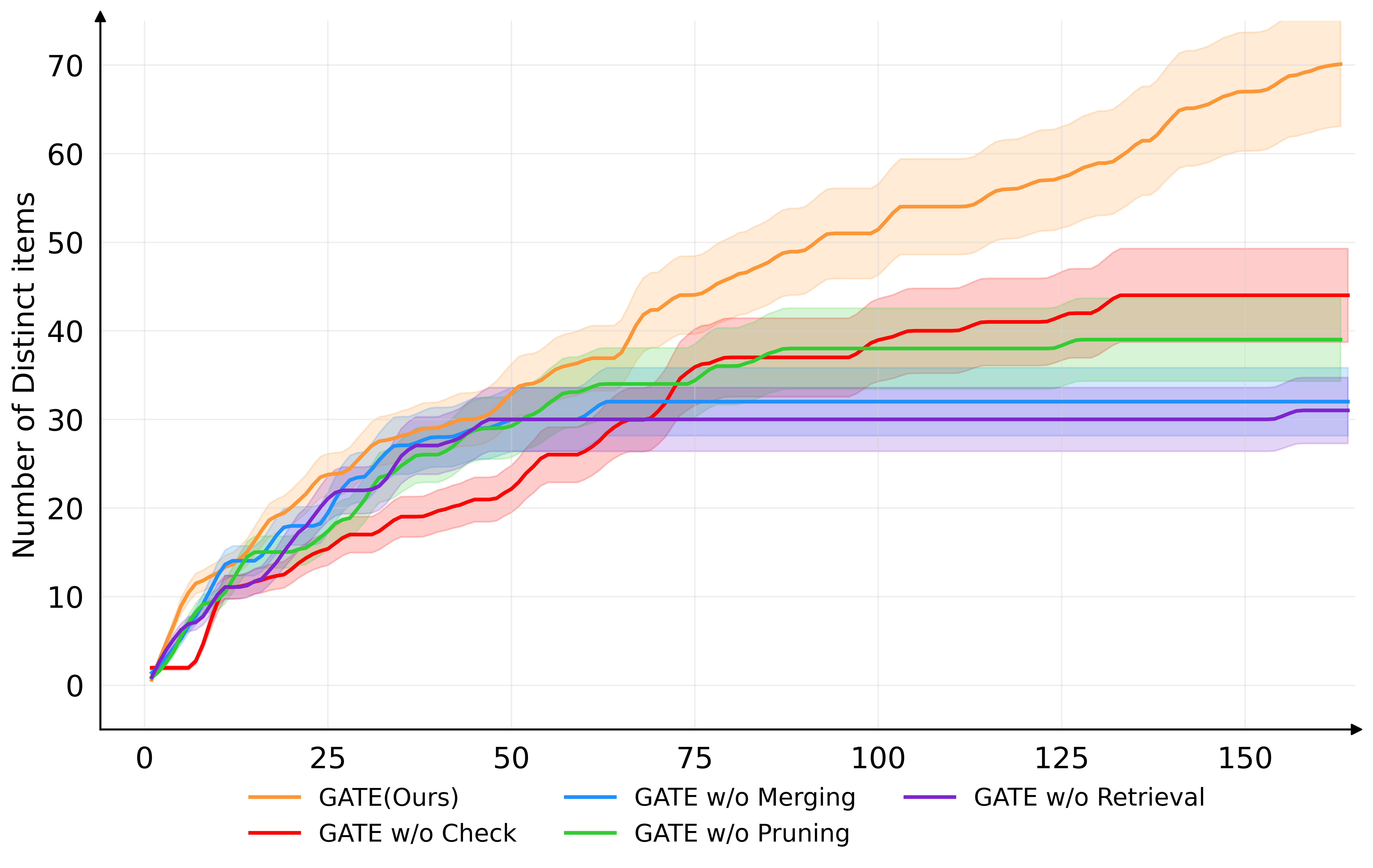}}
\caption{Ablation study of the iterative prompting mechanism. AETN surpasses all other options, highlighting the essential significance of each functional module in the iterative prompting mechanism.}
\label{fig:ablation}
\end{center}
\vskip -0.3in
\end{figure}

As shown in Figure \ref{fig:ablation}, AETG significantly outperforms methods that lack certain functional modules in discovering new Minecraft items and skills. It can be observed that the performance is worst when "w/o retrieval" is used, indicating that the absence of retrieval has the greatest impact on overall functionality and plays a crucial role, thereby validating the effectiveness of our retrieval method. The performance with "w/o duplication" is slightly better, indicating its importance is weaker than that of "w/o retrieval." The performance of "w/o check" and "w/o pruning" is better, but still far behind AETG, which further demonstrates the importance and effectiveness of each functional component.

\vskip -0.1in

\subsection{Closed-Ended Task}
\label{subsec:closed-ended}
For the Closed-Ended Task, we select Textcraft from the Agent Task and Date from the Single-turn Code Task to evaluate the effectiveness of several components in our method. The results are shown in the Table \ref{tab:closed-toolnumber}.

\begingroup
\begin{table}[H]
\caption{The number of tools in Close-Ended Task.}
\label{tab:closed-toolnumber}
\vskip -0.1in
\setlength{\tabcolsep}{10pt} 
\begin{center}
\begin{small}
\begin{sc}
\begin{tabular}{l|cc}
\toprule
\textnormal{\textbf{Method}} & \textnormal{\textbf{TextCraft}}  & \textnormal{\textbf{Date}} \\
\midrule         

\normalfont W/o Self-Check & 42 & 9 \\
\midrule  
\normalfont W/o Merging & 49 & 11\\
\midrule  
\normalfont W/o pruning & 46 & 9 \\
\midrule  
\normalfont GATE & 44 & 4 \\

\bottomrule
\end{tabular}
\end{sc}
\end{small}
\end{center}
\vskip -0.1in
\end{table}
\endgroup

\section{Tool Making}
\label{app:toolgarph}
\subsection{Basic Tools}
\label{subsec:basic-tools}
As shown in the Table \ref{tab:basictool} , the basic tools generated by each method are displayed.

\begingroup
\begin{table}[H]
\caption{Basic tools in various methods.}
\label{tab:basictool}
\vskip -0.1in
\setlength{\tabcolsep}{10pt} 
\begin{center}
\begin{small}
\begin{sc}
\begin{tabular}{l|p{12cm}}
\toprule
\textnormal{\textbf{Tasks}} & \textnormal{\textbf{Basic Tools}}  \\
\midrule         

\normalfont Other Tasks & \normalfont ToolRequest, NotebookBlock, Terminate, CreateTool, EditTool, Python, Feedback, SendAPI, Feedback, Retrieval \\
\midrule  
\normalfont Minecraft & \normalfont smeltItem, killMob, waitForMobRemoved, givePlacedItemBack, useChest, exploreUntil, craftItem, mineBlock, shoot, placeItem, craftHelper, smeltItem, mineflayer, killMob, useChest, exploreUntil, craftItem, mineBlock, placeItem \\

\bottomrule
\end{tabular}
\end{sc}
\end{small}
\end{center}
\vskip -0.1in
\end{table}
\endgroup

\subsection{Tool construction Lists}
\label{subsec:tool construction}

\paragraph{CREATOR:}
\begin{itemize}[noitemsep, topsep=0pt]
    \item \textbf{MATH:}  \textit{sum of areas, find largest won matches, find K, total distance after bounces, find common ratio sum, count lattice points with distance squared, find c for radius, find circle equation and constants, polynomial degree product, calculate cells, find fiftieth term, find non domain values, inverse function product, find m and n, sum of fractions from roots, find roots of quadratic, main, find coefficients, compute expression, prime factors, find x y, find second largest angle, find y coordinate, find constants, evaluate expression, find b for one solution, find c, find minimum value, find possible s, solve expression, find cone height, solve abc, find minimum expression, \dots, time to hit ground, sum of reciprocals of roots, solve x floor x product, sum of possible x, find constant a, sum of squares of solutions, find cost per extra hour, is triangular number, find smallest b greater than 2011, solve exponential equation, solve club suit equation, find degree of h, f, find vertical asymptotes, domain width, maximize revenue, future value, total savings, find min interest rate, equation, find integers, sum of x coordinates squared, find integer values of a, smallest c for real domain, smallest integer c, find m, required investment, simplify expression, g, distance between midpoints, compute x and power, greatest possible a, find continued fraction value, find a b, solve mnp, compute sum, sum of integers in range,
    }

    \item \textbf{Date:}  \textit{get us thanksgiving date, get date one week from first monday of 2019, calculate anniversary date, calculate yesterday from last day of january, calculate one week ago from first monday, get first monday of 2019, calculate yesterday, calculate yesterday from rescheduled meeting, calculate date a month ago from rescheduled meeting, calculate yesterday from first monday of 2019, get date 10 days before us thanksgiving, calculate one week ago from egg runout, calculate one week ago from end of first quarter, calculate date 24 hours later, calculate date a month ago, calculate date 24 hours after anniversary, calculate one week from today from rescheduled meeting, \dots, get tomorrow from us thanksgiving, calculate yesterday from day before yesterday, calculate yesterday from anniversary, calculate date 10 days ago, calculate one year ago from egg run out date, calculate tomorrow from yesterday, calculate one week from last day of january, calculate one week from anniversary, calculate yesterday from eggs run out, calculate tomorrow from today, calculate tomorrow from day before yesterday, calculate one week ago from today, calculate one week ago, calculate date one month ago from anniversary, calculate one year ago from given date, calculate one week from given date}

    \item \textbf{TabMWP:}  \textit{calculate total cost, smallest points, price difference, cost of river rafts, calculate median, calculate range, calculate total spent, rate of change, cost difference, cost for rides, rate of change vacation days, total participants, calculate mean glasses, find mode of states visited, rate of change straight A students, calculate median basketball hoops, count bins with toys in range, people with at least 3 trips, count teams with fewer than 80 swimmers, calculate median clubs, count exact pushups, children with less than 2 necklaces, people played exactly 3 times, count people with fewer than 80 pullups, range of states visited, find spent amount, \dots, calculate median miles, people with fewer than 3 seashells, calculate median glasses, cost to buy cockatiels, largest broken lights, calculate spent, calculate ice cream cost, range of soccer fields, patrons with at least 2 books, count bushes with 20 roses, total people played golf, range of articles, count shipments with exactly 60 broken plates, total cost for lip balms, rate of change scholarships, count teams with fewer than 50 members, count tests with 34 problems, find mode of soccer fields, rate of change hockey games, find lowest score, count pizzas with exactly 48 pepperoni, count people with at least 30 points, cost of wooden benches, rate of change students, patients with fewer than 2 trips, find mode, total cost for hazelnuts, calculate mean fan letters, readers with at least 4 hats, count classrooms with 41 desks}
\end{itemize}

\paragraph{CRAFT:}
\begin{itemize}[noitemsep, topsep=0pt]
    \item  \textbf{MATH:}  \textit{find pack size, count distinct solutions, calculate points, find tank capacity, solve exponential log equation, total energy equilateral triangle, inverse square law force, find max value, total logs in stack, sum of multiples of 13, calculate exponential growth, gravitational force, find x for piecewise composition, positive difference, specific piecewise func, day exceeds 200 cents, find lattice points, count integer parameters for integer solutions, count zeros in square of power of ten minus one, energy stored, sum of squares of roots, sum odd integers, find d minus e squared, compute complex series sum, total energy configuration, sum of areas, \dots, max item price, solve two variable system, inverse variation power, total distance hopped, is prime, total distance, find constant term of polynomial, total distance moved, find perpendicular slope, calculate inverse proportionality, find value of A, count integer a, find min items for higher score, apply r n times, find min x, day exceeds threshold, calculate area in square yards, solve log equation, total items produced, find variable for distance condition, solve time at speeds, find largest solution, find weight of object, calculate proportional value, calculate material cost, solve for variable, total elements in arithmetic sequence, transformed domain, find day for algae coverage, calculate energy stored, least value of y, solve bowling ball weight, find min froods}

    \item \textbf{Date:} \textit{get today date, calculate one week ago, calculate n days from future date, calculate n days from date in format, calculate date days ago, calculate n months from date, calculate one week from today, calculate date after event, find palindrome day, calculate date a month ago, calculate date after days and months, calculate relative date, calculate n days from reference, calculate one year ago from today, calculate n hours from date, calculate date n days from, get date today, calculate date 10 days ago from deadline, calculate n weeks from date, \dots, calculate n units from date, calculate n years from date, calculate n weeks from first weekday of year, calculate today from tomorrow, find special day, calculate date 10 days ago from future, calculate n days after event, calculate date from days passed, calculate one week from christmas eve, calculate one year ago, calculate date 24 hours later, calculate n weeks from anniversary, calculate tomorrow from uk format date, calculate n days from date, is palindrome, calculate one week from first monday of year, calculate one week ago from anniversary}

    \item \textbf{TabMWP:} \textit{get frequency, calculate volleyballs in lockers, calculate total cost from package prices, calculate total items from group counts, calculate mode, calculate donation difference for person, count bags with 20 to 40 broken cookies, calculate total items from groups and items per group, count commutes of 50 minutes, get received amount, calculate total items for groups, find probability, calculate vacation cost, calculate rate of change, find received amount for transaction, calculate vote difference between two items for group, count customers, find minimum value in stem leaf, calculate metric wrenches, find smallest number, count books with 30 to 50 characters, \dots, count people with 67 pullups, calculate difference in donations for person, calculate total cost from unit price and weight, calculate total items from ratio, calculate total cost from unit weight prices and weight, calculate donation difference between causes, calculate difference, calculate net income, calculate grasshoppers on twigs, count total members in group, calculate expenses on date, find lightest child, calculate difference in amounts, count votes for item from groups, calculate probability from count table, get table cell value, calculate jeans in hampers, count instances with specific value in stem leaf, calculate donation difference for person and causes, calculate total from frequency and additional count, calculate range, calculate total reviews}
\end{itemize}

\paragraph{REGAL:}
\begin{itemize}[noitemsep, topsep=0pt]
    \item \textbf{MATH:}  \textit{solve for largest side, apply function sequence, solve rational equation, calculate expression sum, max sum of products, find b for perpendicular bisector, vertex of quadratic, calculate work days, calculate c for zero coefficient, simplify and rationalize sympy, find a for binomial square, compound interest, calculate inverse variation, expand expression, calculate average speed, calculate rs, sum sequence, solve for p, max consecutive integers, find x intercept, day exceeding threshold, find smallest sum, solve for ac pair, constant function, sum of distances, evaluate expression, sum finite geometric series, factor expression, find common difference, total coins pirates, calculate geometric first term, calculate closest whole number, calculate x minus y squared, solve letter values, find circle center v2, evaluate expression with sqrt, calculate sum of equations, \dots, calculate x3 plus y3, find negative intervals, calculate floor and abs, solve quadratic and find min, calculate y, solve for a, check equations, rationalize and simplify, calculate xyz, calculate distance, solve for x in simplified equation, calculate expression, calculate exponent, sum arithmetic series, complete square form, calculate x2 plus y2
    }

    \item \textbf{Date:}  \textit{subtract weeks from date, add weeks to date, format date, add days to date, subtract months from date, subtract days from date, subtract years from date, calculate date, calculate days between weekdays}

    \item \textbf{TabMWP:}  \textit{count range, find mode, total participants, count bushes with fewer roses, find max frequency, total items, count in range, calculate total items, count below threshold, count teams with minimum size, calculate total, calculate range, calculate fraction, sum frequencies below threshold, sum frequencies, calculate difference, calculate median, total outcomes, count specific height, count numbers in range, difference between groups, access frequency, calculate proportionality constant, count values below threshold, find median, calculate probability, calculate mode, get frequency, convert stem leaf to numbers, find minimum, get total items, count scores above, rate of change, calculate mean}
\end{itemize}

\subsection{The tool graph evolution diagrams of {\ours} for various tasks.}
\label{subsec:tool-graph}
Below are the tool graph evolution diagrams for various tasks. The Date task does not have a tool network evolution diagram, as date reasoning does not heavily rely on tool diversity.

\vskip -0.2in

\vskip -0.2in

\section{Prompt Template}
\label{app:prompt}
In this section, we provide the prompt templates of different types used throughout our experiment. These prompts were carefully crafted to ensure that the model's output aligns with the specific objectives of each task.

\subsection{Construction Stage}
In open-ended task online training, we made slight modifications to their prompts based on Voyager~\citep{wang2023voyager}. For close-ended tasks, the prompts used during the construction process are as follows:
\begin{tcolorbox}[title=Task Solver's Prompt, breakable, width=\textwidth,top=0mm]
\begin{Verbatim}[breaklines, fontsize=\footnotesize]
# Instruction #
You are the Task Solver in a collaborative team, specializing in reasoning and Python programming. Your role is to analyze tasks, collaborate with the Tool Manager, and solve problems step by step.
Directly solving tasks without tool analysis is not allowed. Request necessary tools before proceeding when needed, based on the task analysis.

# WORKFLOW #
You can decide which step to take based on the environment and current situation, adapting dynamically as the task progresses.
Stage 1. Tool Requests:
    Requesting tool is mandatory. Request generalized and reusable tools to solve the task. Focus on abstract functionality rather than task-specific details to enhance flexibility and adaptability.
Stage 2. Code and Interact: 
    Write notebook blocks incrementally, executing and interacting with the environment step by step. Avoid bundling all steps into a single block; instead, adjust dynamically based on feedback after each interaction.
Stage 3: Validate and Conclude: 
    When confident in the solution, review your work, validate the results, and conclude the task.

# Custom Library #
===api===

# NOTICE #
1. You must fully understand the action space and its parameters before using it.
2. If code execution fails, you should analyze the error and try to resolve it. If you find that the error is caused by the API, please promptly report the error information to the Tool Manager.
3. Regardless of how simple the issue may seem, you should always aim to summarize and refine the tool requirements.


# Tool Request Guidelines #
1. Keep It Simple: Design tools with single and simple functionality to ensure they are easy to implement, understand, and use. Avoid unnecessary complexity.
2. Define Purpose: Clearly outline the tool’s role within broader workflows. Focus on creating reusable tools that solve abstract problems rather than task-specific ones.
3. Specify Input and Output: Define the required input and expected output formats, prioritizing generic structures (e.g., dictionaries or lists) to enhance flexibility and adaptability.
4. Generalize Functionality: Ensure the tool is not tied to a specific task. Abstract its functionality to make it applicable to similar problems in other contexts.


# ACTION SPACE #
You should Only take One action below in one RESPONSE:
## NotebookBlock Action
* Signature: 
NotebookBlock():
```python
executable python script
```
* Description: The NotebookBlock action allows you to create and execute a Jupyter Notebook cell. The action will add a code block to the notebook with the content wrapped inside the paired ``` symbols. If the block already exists, it can be overwritten based on the specified conditions (e.g., execution errors). Once added or replaced, the block will be executed immediately.
* Restrictions: Only one notebook block can be managed or executed per action.
* Example
- Example1: 
NotebookBlock():
```python
# Calculate the area of a circle with a radius of 5
radius = 5
area = 3.1416 * radius ** 2
print(area)
```

## Tool_request Action
* Signature:
{
    "action_name": "tool_request",
    "argument": {
         "request": [
             ...
         ]
    }
}
* Description: The Tool Request Action allows you to send tool requirements to the Tool Manager and request it to create appropriate tools. You need to provide the action in a JSON format, where the argument field contains a request parameter that accepts a list. Each element in the list is a string describing the desired tool.
* Note:
* Examples:
- Example 1:
{
    "action_name": "tool_request",
    "argument": {
        "request": [
            "I need a tool that calculates the average value of a specified column in a dataset. The input should include the column name."
        ]
    }
}
- Example 2:
{
    "action_name": "tool_request",
    "argument": {
        "request": [
            "I need a tool that filters rows in a dataset based on a specified condition. The input should include the column name and the condition to filter by."
        ]
    }
}


## Terminate Action
* Signature: Terminate(result=the result of the task)
* Description: The Terminate action ends the process and provides the task result. The `result` argument contains the outcome or status of task completion.
* Examples:
  - Example1: Terminate(result="A")
  - Example2: Terminate(result="1.23")

# RESPONSE FORMAT #
For each task input, your response should contain:
1. One RESPONSE should only contain One Stage, One Thought and One Action.
2. An current phase of task completion, outlining the steps from planning to review, ensuring progress and adherence to the workflow.  (prefix "Stage: ").
3. An analysis of the task and the current environment, including reasoning to determine the next action based on your role as a SolvingAgent. (prefix "Thought: ").
4. An action from the **ACTION SPACE** (prefix "Action: "). Specify the action and its parameters for this step.

# RESPONSE EXAMPLE #
Observation: ...(the output of last actions, as provided by the environment and the code output, you don't need to generate it)

Stage:...(One Stage from `WORKFLOW`)
Thought: ...
Action: ...(Use an action from the ACTION SPACE no more than once per response.)

# TASK #
===task===
\end{Verbatim}
\end{tcolorbox}

\begin{tcolorbox}[title=Tool Manager's Prompt, breakable, width=\textwidth,top=0mm]
\begin{Verbatim}[breaklines, fontsize=\footnotesize]
# Instruction #
You are a Tool Manager in a collaborative team, specializing in assembling existing APIs to construct hierarchical and reusable abstract tools based on predefined criteria.
You will be provided with a custom library, similar to Python’s built-in modules, containing various functions related to date reasoning. For each task, you will receive:
1. Tool request: The specific goal or functionality the new tool must achieve.
2. Existing tools: A list of available functions from the custom library that you can utilize.
Your task is to analyze the given request and create a reusable tool by effectively leveraging the relevant functions from the existing tools or utilizing basic tools to achieve the desired functionality. 
If an existing tool from the provided library already fully satisfies the requirements, simply return that tool instead of duplicating functionality. Ensure all responses align with reusability and efficiency principles.

# Custom Library #
===api===

# Creation Criteria #
- **Reusability**: The function could be resued for more complex function.
- **Innovation**: Tools should offer innovation, not merely wrap or replicate existing APIs. Simply re-calling an API without significant enhancements does not qualify as innovation.
- **Completeness**: The function should handle potential edge cases to ensure completeness.
- **Leveraging Existing Functions**: The function should effectively utilize existing functions to enhance efficiency and avoid redundancy.
- **Functionality**: Ensure the tool runs successfully and is bug-free, guaranteeing full functionality.

# ACTION SPACE #
You should Only take One action below in one RESPONSE:
## Create tool Action
* Description: The Create Tool action allows you to develop a new tool and temporarily store it in a private repository accessible only to you. Each invocation creates a single tool at a time. You can repeatedly use this action to build smaller components, which can later be assembled into the final tool.
* Signature: 
Create_tool(tool_name=The name of the tool you want to create):
```python
The source code of tool
```
* Example:
Create_tool(tool_name=“calculate_column_statistics”):
```python
def calculate_column_statistics(dataset: pd.DataFrame, column_name: str) -> Dict[str, float]:
    """
    Calculates basic statistics (mean, median, standard deviation) for a specified column in a dataset.
    Parameters:
    - dataset: A pandas DataFrame containing the data.
    - column_name: The name of the column to calculate statistics for.
    Returns:
    - A dictionary containing the mean, median, and standard deviation of the column.
    """
    if column_name not in dataset.columns:
        raise ValueError(f"Column '{column_name}' not found in the dataset.")
    
    column_data = dataset[column_name]
    stats = {
        "mean": column_data.mean(),
        "median": column_data.median(),
        "std_dev": column_data.std()
    }
    return stats
```
## Edit tool Action
* Description: The Edit Tool action allows you to modify an existing tool and temporarily store it in a private repository that only you can access. You must provide the name of the tool to be updated along with the complete, revised code. Please note that only one tool can be edited at a time.
* Signature: 
Edit_tool(tool_name=The name of the tool you want to create):
```python
The edited source code of tool
```
* Examples:
Edit_tool(tool_name="filter_rows_by_condition"):
```python
def filter_rows_by_condition(dataset: pd.DataFrame, column_name: str, condition: str) -> pd.DataFrame:
    """
    Filters rows in a dataset based on a specified condition for a given column.
    Parameters:
    - dataset: A pandas DataFrame containing the data.
    - column_name: The name of the column to apply the condition to.
    - condition: A string representing the condition, e.g., 'value > 10'.
    Returns:
    - A filtered DataFrame containing only the rows that satisfy the condition.
    """
    if column_name not in dataset.columns:
        raise ValueError(f"Column '{column_name}' not found in the dataset.")
    
    try:
        filtered_dataset = dataset.query(f"{column_name} {condition}")
    except Exception as e:
        raise ValueError(f"Invalid condition: {condition}. Error: {e}")
    
    return filtered_dataset
```

# RESPONSE FORMAT #
For each task input, your response should contain:
1. Each response should contain only one "Thought," and one "Action."
2. Determine how to construct your tool to meet tool request and function creation criteria. Check if any functions in the Existing Tool can be invoked to assist in the tool’s development and ensure alignment with the criteria.(prefix "Thought: ").
3. An action dict from the **ACTION SPACE** (prefix "Action: "). Specify the action and its parameters for this step. 

# RESPONSE EXAMPLE  #
1. If you determine that the tool request cannot be solved using existing tools, choose this mode to provide a clear and complete code solution.

Thought: ...
Action: ...

2. If you determine that the tool request is already satisfied by existing tools, choose this mode to directly reference and return the relevant tool without creating additional solutions.
Thought: ...
Tool: {  
    "tool_name": "Name of Existing tools"
}

# NOTICE #
1. You can directly call and use the tool in the custom library in your code or tool without importing it.
2. You can only create or edit one tool per response, so take it one step at a time.

# TASK #
===task===
\end{Verbatim}
\end{tcolorbox}

\begin{tcolorbox}[title=Prompt of Self-Check Step 1, breakable, width=\textwidth,top=0mm]
\begin{Verbatim}[breaklines, fontsize=\footnotesize]
# Instruction #
You are evaluating whether the tools provided by the Tool Manager meet the required standards. 
You follow a defined workflow, take actions from the ACTION SPACE, and apply the evaluation criteria. 

# Evaluation Criteria #
- **Reusability**: The function should be designed for reuse in more complex scenarios. For instance, in the case of the `craft_wooden_sword()` tool, it would be more versatile if it could accept a quantity as an input parameter.
- **Innovation**: Tools should offer innovation, not merely wrap or replicate existing APIs. Simply re-calling an API without significant enhancements does not qualify as innovation. If an existing tool from the provided library already fully satisfies the requirements, simply return that tool instead of duplicating functionality. Ensure all responses align with reusability and efficiency principles.
- **Completeness**: The function should handle potential edge cases to ensure completeness.
- **Leveraging Existing Functions**: Check if any function in "Existing Function" is helpful for completing the task. If such functions exist but are not invoked in the provided code, relevant feedback should be given.

## Tool Abstraction ##
Tool abstraction is essential for enabling tools to adapt to diverse tasks. Key principles include:
- Design generic functions to handle queries of the same type, based on shared reasoning steps, avoiding specific object names or terms.
- Name functions and write docstrings to reflect the core reasoning pattern and data organization, without referencing specific objects.
- Use general variable names and pass all column names as arguments to enhance adaptability.

# ACTION SPACE #
You should Only take One action below in one RESPONSE:
# Feedback Action
* Signature: {
    "action_name": "Feedback",
    "argument": {
        "feedback": ...
        "passed": true/false
    }
}
* Description: The Feedback Action is represented as a JSON string that provides feedback to the Tool Manager or SolvingAgent. The feedback field contains comments or suggestions, while pass indicates whether the tool meets the requirements (true for approval, false for rejection). Feedback should be concise, constructive, and relevant. If pass is true, the feedback can be left empty; otherwise, it must be provided.
* Example:
- Example1:
{
    "action_name": "Feedback",
    "argument": {
        "feedback": "",
        "passed": true
    }
}
- Example2:
{
    "action_name": "Feedback",
    "argument": {
        "feedback": "The tool correctly solves the equation for small numbers, but fails when the coefficients are very large. Consider optimizing the algorithm for handling larger values and improving computational efficiency.",
        "passed": false
    }
}

# RESPONSE FORMAT #
For each task input, your response should contain:
1. One RESPONSE should ONLY contain One Thought and One Action.
2. An comprehensive analysis of the tool code based on the evaluation criteria.(prefix "Thought: ").
3. An action from the **ACTION SPACE** (prefix "Action: "). 

# EXAMPLE RESPONSE #
Observation: ...(output from the last action, provided by the environment and task input, no need for you to generate it)

Thought: 1. Reusability: ...
2. Innovation: ...
3. Completeness: ...
4. Leveraging Existing Functions: ...

Action: ...(Use an action from the ACTION SPACE once per response.)

# Custom Library #
===api===

# TASK #
===task===
\end{Verbatim}
\end{tcolorbox}

\begin{tcolorbox}[title=Prompt of Self-Check Step 2, breakable, width=\textwidth,top=0mm]
\begin{Verbatim}[breaklines, fontsize=\footnotesize]
# Instruction #
You are verifying whether the tools provided by the Tool Manager execute without runtime errors.
You will use a custom library, similar to the built-in library, which provides everything necessary for the tasks. Your task is only to execute the provided tool code and check for runtime errors, not to evaluate the tool’s functionality or correctness.

# Stage and Workflow #
1. **Ensure Bug-Free Tool Operation**:
	- Execute the tool to ensure it runs without any runtime bugs.
	- You don’t need to verify the function’s functionality; simply call it to check for any runtime errors.
	- If the tool is a retrieved API, skip this step and proceed.
2. **Send Feedback**:
	- After executing the code, provide feedback based on the output, indicating whether the operation was successful or not.

# Notice #
1. If any issues with the tool are found, promptly provide clear and critical feedback to the Tool Manager for resolution. 
2. You should not create or edit functions (tools) with the same name as the Existing Functions in the code.
3. You can directly call the APIs from the custom library without needing to import or declare any external libraries.
4. You don’t need to verify the function’s functionality or set up its standard output; simply call it to check for any errors.

# ACTION SPACE #
You should Only take One action below in one RESPONSE:
## Python Action
* Signature: 
Python(file_path=python_file):
```python
executable_python_code
```
* Description: The Python action will create a python file in the field `file_path` with the content wrapped by paired ``` symbols. If the file already exists, it will be overwritten. After creating the file, the python file will be executed. Remember You can only create one python file.
* Examples:
- Example1
Python(file_path="solution.py"):
```python
# Calculate the area of a circle with a radius of 5
radius = 5
area = 3.1416 * radius ** 2
print(f"The area of the circle is {area} square units.")
```
- Example2
Python(file_path="solution.py"):
```python
# Calculate the perimeter of a rectangle with length 8 and width 3
length = 8
width = 3
perimeter = 2 * (length + width)
print(f"The perimeter of the rectangle is {perimeter} units.")
```

# Feedback Action
* Signature: {
    "action_name": "Feedback",
    "argument": {
        "feedback": ...
        "passed": true/false
    }
}
* Description: The Feedback Action is used to provide feedback to the Tool Manager. The feedback field contains detailed comments or suggestions. If the tool encounters an error, you should set passed to false and provide a detailed feedback. If the tool runs without errors, you can set passed to true and leave feedback as an empty string.
* Examples:
- Example 1:
{
    "action_name": "Feedback",
    "argument": {
        "feedback": ""
        "passed": true
    }
}
- Example 2:
{
    "action_name": "Feedback",
    "argument": {
        "feedback": "The tool encountered an error while executing. The variable 'height' is missing in the function call. Please ensure that all required parameters are provided.",
        "passed": false
    }
}

# RESPONSE FORMAT #
For each task input, your response should contain:
1. One RESPONSE should ONLY contain One Thought and One Action.
2. An analysis of the task and current environment, reasoning through the next evaluation step based on your role as CheckingAgent.(prefix "Thought: ").
3. An action from the **ACTION SPACE** (prefix "Action: "). Specify the action and its parameters for this step.

# EXAMPLE RESPONSE #
Observation: ...(output from the last action, provided by the environment and task input, no need for you to generate it)

Thought: ...
Action: ...(Use an action from the ACTION SPACE once per response.)

# Custom Library #
You can use pandas, sklearn, or other Python libraries as part of the custom library.

* Note: You can directly call these tools without importing or redefining them in your code.

Let's think step by step.
# TASK #
===task===
\end{Verbatim}
\end{tcolorbox}

\subsection{Test Stage}
\label{appsub:test_prompt}
During the test stage, the prompts used for different datasets are as follows:
\begin{tcolorbox}[title=Prompt on DABench, breakable, width=\textwidth,top=0mm]
\begin{Verbatim}[breaklines, fontsize=\footnotesize]
# Instruction #
You are a helpful assistant, skilled in data science tasks.
You will be provided with a task description and related files. 
You should complete tasks by writing notebook code to interact with the environment containing the task files.
Additionally, you must strictly adhere to the task constraints. 
Once the task is completed, you need to format the answer as specified in the answer format and invoke the Terminate action to conclude.
You should use actions from the ACTION SPACE, follow the Response Format, and complete the task within 20 steps.

You may also leverage the following helper functions if needed.
===api===


===example===


# Response Format #
Your each response should contain:
1. One RESPONSE should only contain ONLY One Thought and ONLY One Action.
2. Only an analysis of the task and the current environment, including reasoning to determine the next action. (prefix "Thought: ").
3. Only an action from the **ACTION SPACE** (prefix "Action: "). Specify the action and its parameters for this step.

Observation: ...(Provided by the environment, no need for you to generate it.))

Thought: ...
Action: ...

# ACTION SPACE #
## NotebookBlock Action
* Signature: 
NotebookBlock():
```python
executable python script
```
* Description: The NotebookBlock action allows you to create and execute a Jupyter Notebook cell. The action will add a code block to the notebook with the content wrapped inside the paired ``` symbols. If the block already exists, it can be overwritten based on the specified conditions (e.g., execution errors). Once added or replaced, the block will be executed immediately.
* Restrictions: Each response must contain only one notebook block.
* Note: In a single block, you may call multiple tools or single.
* Example:
Action: NotebookBlock():
```python
# Calculate the area of a circle with a radius of 5
radius = 5
area = 3.1416 * radius ** 2
print(area)
```

# Terminate Action
* Signature: Terminate(result="the result of the task")
* Description: The Terminate action marks the completion of a task and presents the final result. It is a formatting guideline, not an executable Python function. The result parameter must contain a clear, specific answer that strictly complies with the task’s output format, with all required values explicitly provided.
Tips:
    - Ensure the result parameter provides a definite and concrete final answer.
    - Do not include unresolved Python expressions, placeholders, or variables (e.g., @value[{x + y}] or @result[{variable_name}] or "@result[{variable_name}]".format(variable_name)).
    - The output must adhere precisely to the task’s formatting specifications, ensuring clarity and consistency.
* Examples:
- Example 1: 
Answer Format: @shapiro_wilk_statistic[test_statistic] @shapiro_wilk_p_value[p_value]
Action: Terminate(result="@shapiro_wilk_statistic[0.56] @shapiro_wilk_p_value[0.0002]")
- Example 2: 
Answer Format: @total_votes_outliers_num[outlier_num]
where "outlier_num" is an integer representing the number of values considered outliers in the 'total_votes' column.
Action: Terminate(result="@total_votes_outliers[10]")
- Example3:
Action: Terminate(result="@normality_test_result[Not Normal] @p_value[0.000]")

## Response Example
Here are four examples of responses.
## Example1
Thought: The dataset has been loaded successfully and it contains the "Close Price" column. Now, we need to calculate the mean of the "Close Price" column using pandas.
Action: NotebookBlock():
```python
# Calculate the mean of the "Close Price" column
mean_close_price = data["Close Price"].mean()
# Round the result to two decimal places
mean_close_price_rounded = round(mean_close_price, 2)
print(mean_close_price_rounded)
```
## Example2
Thought: We need to filter the dataset to only include rows where the “Volume” is greater than 100,000. This will help focus on high-volume trades.
Action: NotebookBlock():
```python
# Filter rows where "Volume" is greater than 100,000
filtered_data = data[data["Volume"] > 100000]
# Display the filtered dataset
print(filtered_data)
```
## Example3
Thought: To analyze the correlation between “Open Price” and “Close Price,” we will calculate the Pearson correlation coefficient using pandas.
Action: NotebookBlock():
```python
# Calculate the correlation between "Open Price" and "Close Price"
correlation = data["Open Price"].corr(data["Close Price"])
# Print the correlation result
print(correlation)
```
## Example4
Thought: To check for missing values in the dataset, we need to check for null values in each column using pandas.
Action: NotebookBlock():
```python
# Check for missing values in each column
missing_values = data.isnull().sum()
# Display the result
print(missing_values)
```

# Begin #
Let's Begin.
## Task 
===task===
\end{Verbatim}
\end{tcolorbox}

\begin{tcolorbox}[title=Prompt on TextCraft, breakable, width=\textwidth,top=0mm]
\begin{Verbatim}[breaklines, fontsize=\footnotesize]
# Instruction #
You are provided with a set of useful crafting recipes to create items in Minecraft.
Crafting commands follow the format: "craft [target object] using [input ingredients]".
You can either "fetch" an object (ingredient) from the inventory or the environment or "craft" the target object using the provided crafting commands.
You are allowed to use only the crafting commands provided; do not invent or use your own crafting commands.
If a crafting command specifies a generic ingredient, such as "planks", you can substitute it with a specific type of that ingredient (e.g., “dark oak planks”).
To complete the crafting tasks, you will write notebook code utilizing tools from the "Custom Library". You should carefully read and understand the tool’s docstrings and code to fully grasp their functionality and usage.
The tools should be invoked by outputting a block of Python code. Additionally, you may incorporate Python constructs such as for-loops, if-statements, and other logic where necessary.
Please always use actions from the ACTION SPACE and follow the Response Format.


# ACTION SPACE #
## NotebookBlock Action
* Signature: 
NotebookBlock():
```python
executable python script
```
* Description: The NotebookBlock action creates and executes a Jupyter Notebook cell. It adds a code block wrapped in ``` symbols, overwriting existing blocks if specified (e.g., after execution errors). The block is executed immediately after being added or replaced.
* Note: In a single block, you may call multiple tools.

## Terminate Action
* Signature: Terminate(result=the result of the task)
* Description: The Terminate action ends the process and provides the task result. The `result` argument contains the outcome or status of task completion. Only the CheckingAgent has the authority to decide whether a task is finished.
* Examples:
  - Example1: Action: Terminate(result="3")
  - Example2: Action: Terminate(result="Successfully craft 2 oak planks")
  - Example3: Action: Terminate(result="Successfully craft 1 milk")


# Response Format #
For each task input, your response should contain:
1. One RESPONSE should only contain ONLY One Thought and ONLY One Action.
2. An analysis of the task and the current environment, including reasoning to determine the next action. (prefix "Thought: ").
3. An action from the **ACTION SPACE** (prefix "Action: "). Specify the action and its parameters for this step.

## Response Example
### Example1
Thought: To craft a crimson fence, I need to use the crafting command: “craft 3 crimson fence using 2 stick, 4 crimson planks.” This means I need two ingredients: 2 sticks and 4 crimson planks. To obtain 4 crimson planks, I can use the command “craft 4 crimson planks using 1 crimson stems,” so I need to fetch 1 crimson stem first. For the sticks, I can use the command “craft 4 stick using 2 planks.” Since the crafting of sticks requires planks, I can directly use the crimson planks I craft earlier for this purpose.
Steps: 
1. Fetch 1 crimson stems.
2. Craft 4 crimson planks using 1 crimson stems.
3. Craft 4 sticks using 2 crimson planks.
4. Craft 3 crimson fences using 2 sticks and 4 crimson planks.
Action: NotebookBlock()
```python
get_object('1 crimson stems')
```
### Example2
Thought: The inventory is currently empty, which means I need to retrieve the necessary spruce logs from the environment. Since we need spruce planks to craft the spruce trapdoor, and each spruce trapdoor requires 6 spruce planks, I should start by fetching an adequate number of spruce logs. Since each spruce log can craft 4 spruce planks, we need at least 2 spruce logs to ensure we can craft 8 spruce planks (more than enough for 2 spruce trapdoors if needed in the future).
Action: NotebookBlock()
```python
# Step 2: Fetch 2 spruce logs from the environment
get_object('2 spruce logs')
```
### Example3
Thought: We have successfully retrieved 2 spruce logs from the environment. 
Action: Terminate(result="Successfully crafted 8 spruce planks")


# Custom Library #
### Tool `check_inventory`
Source Code:
```python
def check_inventory() -> str:
    """
    Retrieves the current inventory state from the environment.
    The function sends an 'inventory' command to the environment
    and processes the observation to return a string representation
    of the inventory, listing items and their quantities.
    Returns:
        str: A string describing the inventory in the format:
             "Inventory: [item_name] (quantity)"
    """
    obs, _ = step('inventory')
    return obs
```
Usage Example:
```python
check_inventory() 
# If the environment has no items, Output: Inventory: You are not carrying anything.
# If the environment contains 2 oak planks, Output: Inventory: [oak planks] (2)
```
### Tool `get_object`
Source Code:
```python
def get_object(target: str) -> None:
    """
    Retrieves an item from the environment.

    The function prints the response message from the environment, 
    indicating whether the retrieval was successful or not.

    Args:
        target (str): The name of the item to be retrieved.

    Returns:
        None
    """
    obs, _ = step("get " + target)
    print(obs)
```
Usage Example:
Craft Command:
craft 2 yellow dye using 1 sunflower
craft 8 yellow carpet using 8 white carpet, 1 yellow dye
```python
get_object("1 sunflower") # Ouput: Got 1 sunflower
get_object("2 sunflower") # Ouput: Got 2 sunflower
# Note: You cannot retrieve yellow dye directly from the environment; it must first be crafted using sunflowers.
get_object("1 yellow dye") # Output: Could not find yellow dye
```
### Tool `craft_object`
Source Code:
```python
def craft_object(target: str, ingredients: List[str]) -> None:
    """
    Crafts a specified item using the given ingredients.

    This function's `target` and `ingredients` parameters correspond to the craft command: 
    "Craft 'target' using [ingredients]".
    
    **Note:** The `ingredients` must exactly match the command format. For example, if the command requires 
    '1 oak logs', providing '1 oak log' instead will not be recognized.

    Prints the environment's response to indicate whether the crafting operation was successful or not.

    Args:
        target (str): The item to craft along with its quantity (e.g., '4 oak planks').
        ingredients (List[str]): A list of required ingredients with their respective quantities 
                                (e.g., ['1 oak logs']).

    Returns:
        None

    """
    obs, _ = step("craft " + target + " using " + ", ".join(ingredients))
    print(obs)
```
Usage Example:
- Example1
Query: Craft 1 black dye using 1 wither rose
Craft Command:
craft 1 black dye using 1 wither rose
craft 1 black dye using 1 ink sac

```python
get_object('1 wither rose')
craft_object('1 black dye', ['1 wither rose'])
# Ouput:
# Got 1 wither rose
# Crafted 1 minecraft:black dye
```
- Example2
Craft Commands: 4 oak planks using 1 oak logs
```python
get_object("1 oak logs") # Got 1 oak logs
check_inventory() # Inventory: ["oak logs"] (1)
craft_object('4 oak planks', ['1 oak logs'])
# Crafted 4 minecraft:oak planks
```

===api===


===example===

# Example #
Here is a demo of how to fetch and craft objects.
Crafting commands:
craft 3 dark oak sign using 6 dark oak planks, 1 stick
craft 4 dark oak planks using 1 dark oak logs
craft 1 stick using 1 planks
craft 4 stick using 2 bamboo
craft 4 oak planks using 1 oak logs
craft 1 dark oak fence using 2 stick, 4 dark oak planks
craft 1 warped stairs using 6 warped planks
craft 3 oak sign using 6 oak planks, 1 stick

craft 2 oak planks

Thought: To craft 2 oak planks, I need to first check if I can fetch them directly from the environment or my inventory. If they are not available, I will craft them using the appropriate crafting recipe. Based on the crafting commands, I know that I can craft 4 oak planks using 1 oak logs. My plan is as follows:
1. Check if 2 oak planks are already in my inventory or can be fetched from the environment.
2. If not, verify if I have the required ingredient (1 oak logs).
3. Gather the missing ingredient, if necessary.
4. Craft 4 oak planks and ensure 2 are available for the task.
Action: NotebookBlock():
```python
check_inventory()
```

Observation: Inventory: [stick] (1) 

Thought: I could not find 2 oak planks or the required ingredient (1 oak logs) in my inventory. I need to fetch 1 oak logs first.
Action: NotebookBlock():
```
get_object("1 oak logs")
```

Observation: Got 1 oak logs

Thought: I now have the required ingredient to craft oak planks. I will use the crafting command: craft 4 oak planks using 1 oak logs.
Action: NotebookBlock():
```
craft_object("4 oak planks", ["1 oak logs"])
check_inventory()
```

Observation: Crafted 4 minecraft:oak_planks
Inventory: [stick] (1) [oak_planks] (4)

Thought: I now have 2 oak planks in my inventory. Task Completed!
Action: Terminate(result=“Successfully craft 2 oak planks”)


# Begin #
Let's Begin. Please follow the Instruction and focus on solve the task.
# Current items in the inventory #
These are the items currently in your inventory. You cannot use items that are not in your inventory to craft new items.
===inventory===

# Task #
===task===
\end{Verbatim}
\end{tcolorbox}

\begin{tcolorbox}[title=Prompt on MATH, breakable, width=\textwidth,top=0mm]
\begin{Verbatim}[breaklines, fontsize=\footnotesize]
Your task is to solve math competition problems by writing Python programs.

You may also leverage the following helper functions, but must avoid fabricating and calling undefined function names.
```python
===api===
```

Examples: 

Examples: 
Query: Point $P$ lies on the line $x= -3$ and is 10 units from the point $(5,2)$. Find the product of all possible $y$-coordinates that satisfy the given conditions.
Program: 
```python
from sympy import symbols, Eq, solve
# Define symbolic variable for y-coordinate of point P
y = symbols('y')
# Step 1: Given conditions
x1 = -3  # Point P lies on the vertical line x = -3
x2, y2 = 5, 2  # Coordinates of the given point (5, 2)
d = 10  # Distance between point P and (5,2)
# Step 2: Apply the distance formula
# Distance formula: sqrt((x2 - x1)^2 + (y - y2)^2) = d
# Squaring both sides to eliminate the square root:
# (x2 - x1)^2 + (y - y2)^2 = d^2
distance_equation = Eq((x2 - x1)**2 + (y - y2)**2, d**2)
# Step 3: Solve for possible values of y
y_solutions = solve(distance_equation, y)
# Step 4: Compute the product of all possible y-values
product = y_solutions[0] * y_solutions[1]
# Step 5: Output the final result
print("Final Answer:", product)
```

Query: If $3p+4q=8$ and $4p+3q=13$, what is $q$ equal to?
Program:
```python
from sympy import symbols, Eq, solve
# Define symbolic variables for the unknowns p and q
p, q = symbols('p q')
# Step 1: Define the given system of equations
eq1 = Eq(3 * p + 4 * q, 8)  # Equation 1: 3p + 4q = 8
eq2 = Eq(4 * p + 3 * q, 13)  # Equation 2: 4p + 3q = 13
# Step 2: Solve the system of equations for p and q
solution = solve((eq1, eq2), (p, q))
# Step 3: Extract and output the value of q
print("Final Answer:", solution[q])
```

Query: Simplify $\frac{3^4+3^2}{3^3-3}$. Express your answer as a common fraction.
Program:
```python
from sympy import symbols, simplify
# Define the variable
x = symbols('x')
# Define the expression
numerator = 3**4 + 3**2
denominator = 3**3 - 3
fraction = numerator / denominator
# Simplify the fraction
simplified_fraction = simplify(fraction)
# Print the result
print("Final Answer:", simplified_fraction)
```

===example===

## Begin !
Please generate ONLY the code wrapped in ```python...``` to solve the query below.

Query: ===task===
Program:
\end{Verbatim}
\end{tcolorbox}

\begin{tcolorbox}[title=Prompt on Date, breakable, width=\textwidth,top=0mm]
\begin{Verbatim}[breaklines, fontsize=\footnotesize]
Your task is to solve simple word problems by creating Python programs.

You may also leverage the following helper functions, but must avoid fabricating and calling undefined function names, such as `calculate_date_by_years`.
```python
===api===
```

Examples:

Query: In the US, Thanksgiving is on the fourth Thursday of November. Today is the US Thanksgiving of 2001. What is the date one week from today in MM/DD/YYYY?
Program:
```python
# import relevant packages
from datetime import date, time, datetime
from dateutil.relativedelta import relativedelta
import calendar
# 1. What is the date of the first Thursday of November? (independent, support: [])
date_1st_thu = date(2001,11,1)
while date_1st_thu.weekday() != calendar.THURSDAY:
    date_1st_thu += relativedelta(days=1)
# 2. How many days are there in a week? (independent, support: ["External knowledge: There are 7 days in a week."])
n_days_of_a_week = 7
# 3. What is the date today? (depends on 1 and 2, support: ["Today is the US Thanksgiving of 2001", "Thanksgiving is on the fourth Thursday of November"])
days_from_1st_to_4th_thu = (4-1) * n_days_of_a_week
date_today = date_1st_thu + relativedelta(days=days_from_1st_to_4th_thu)
# 4. What is the date one week from today? (depends on 3, support: [])
date_1week_from_today = date_today + relativedelta(weeks=1)
# 5. Final Answer: What is the date one week from today in MM/DD/YYYY? (depends on 4, support: [])
answer = date_1week_from_today.strftime("%m/%d/%Y")
# print the answer
print(answer)
```

Query: Yesterday was 12/31/1929. Today could not be 12/32/1929 because December has only 31 days. What is the date tomorrow in MM/DD/YYYY?
Program:
```python
# import relevant packages
from datetime import date, time, datetime
from dateutil.relativedelta import relativedelta
# 1. What is the date yesterday? (independent, support: ["Yesterday was 12/31/1929"])
date_yesterday = date(1929,12,31)
# 2. What is the date today? (depends on 1, support: ["Today could not be 12/32/1929 because December has only 31 days"])
date_today = date_yesterday + relativedelta(days=1)
# 3. What is the date tomorrow? (depends on 2, support: [])
date_tomorrow = date_today + relativedelta(days=1)
# 4. Final Answer: What is the date tomorrow in MM/DD/YYYY? (depends on 3, support: [])
answer = date_tomorrow.strftime("%m/%d/%Y")
# print the answer
print(answer)
```

Query: The day before yesterday was 11/23/1933. What is the date one week from today in MM/DD/YYYY?
Program:
```python
# import relevant packages
from datetime import date, time, datetime
from dateutil.relativedelta import relativedelta
# 1. What is the date the day before yesterday? (independent, support: ["The day before yesterday was 11/23/1933"])
date_day_before_yesterday = date(1933,11,23)
# 2. What is the date today? (depends on 1, support: [])
date_today = date_day_before_yesterday + relativedelta(days=2)
# 3. What is the date one week from today? (depends on 2, support: [])
date_1week_from_today = date_today + relativedelta(weeks=1)
# 4. Final Answer: What is the date one week from today in MM/DD/YYYY? (depends on 3, support: [])
answer = date_1week_from_today.strftime("%m/%d/%Y")
# print the answer
print(answer)
```

===example===

## Begin !
Please generate ONLY the code wrapped in ```python...``` to solve the query below.

Query: ===task===
Program:
\end{Verbatim}
\end{tcolorbox}

\begin{tcolorbox}[title=Prompt on TabMWP, breakable, width=\textwidth,top=0mm]
\begin{Verbatim}[breaklines, fontsize=\footnotesize]
Your task is to solve table-reasoning problems by writing Python programs.
You are given a table. The first row is the name for each column. Each column is seperated by "|" and each row is seperated by "\n".
Pay attention to the format of the table, and what the question asks.

You may also leverage the following helper functions, but must avoid fabricating and calling undefined function names.
```python
===api===
```


Examples: 
### Table
Name: None
Unit: $
Content:
Date | Description | Received | Expenses | Available Funds
 | Balance: end of July | | | $260.85
8/15 | tote bag | | $6.50 | $254.35
8/16 | farmers market | | $23.40 | $230.95
8/22 | paycheck | $58.65 | | $289.60
### Question
This is Akira's complete financial record for August. How much money did Akira receive on August 22?
### Solution code
```python
records = {
    "7/31": {"Description": "Balance: end of July", "Received": "", "Expenses": "", "Available Funds": 260.85},
    "8/15": {"Description": "tote bag", "Received": "", "Expenses": 6.5, "Available Funds": ""},
    "8/16": {"Description": "farmers market", "Received": "", "Expenses": 23.4, "Available Funds": ""},
    "8/22": {"Description": "paycheck", "Received": 58.65, "Expenses": "", "Available Funds": ""}
}
# Access the amount received on August 22
received_aug_22 = records["8/22"]["Received"]
print("Final Answer: ", received_aug_22)
```

### Table
Name: Orange candies per bag
Unit: bags
Content:
Stem | Leaf 
2 | 2, 3, 9
3 | 
4 | 
5 | 0, 6, 7, 9
6 | 0
7 | 1, 3, 9
8 | 5
### Question
A candy dispenser put various numbers of orange candies into bags. How many bags had at least 32 orange candies?
### Solution code
```python
data = {
    2: [2, 3, 9],
    3: [],
    4: [],
    5: [0, 6, 7, 9],
    6: [0],
    7: [1, 3, 9],
    8: [5]
}
# Initialize the count to zero
count = 0
# Iterate over the keys in the dictionary
for key in data:
    # Combine tenth digit and unit digit
    if key * 10 + data[key] >= 32:
        # Increment the count
        count += 1

# Output the result
print("Final Answer: ", count)
```

### Table
Name: Monthly Savings  
Unit: $  
Content:  
Date  | Description       | Received | Expenses | Available Funds |
       | Balance: end of May |   |   | $500.00 |
6/10  | groceries        |   | $45.75 | $454.25 |
6/15  | gas refill       |   | $30.20 | $424.05 |
6/25  | salary           | $1200.00 |   | $1624.05 |
### Question
How much money did Akira receive on June 25?
### Solution code
```python
import pandas as pd
records = {
    "5/31": {"Description": "Balance: end of May", "Received": "", "Expenses": "", "Available Funds": 500.00},
    "6/10": {"Description": "groceries", "Received": "", "Expenses": 45.75, "Available Funds": ""},
    "6/15": {"Description": "gas refill", "Received": "", "Expenses": 30.2, "Available Funds": ""},
    "6/25": {"Description": "salary", "Received": 1200.00, "Expenses": "", "Available Funds": ""}
}
# Access the amount received on June 25
received_june_25 = records["6/25"]["Received"]
print("Final Answer: ", received_june_25)
```

===example===

## Begin!
Please solve the task below and enclose your code within a single code block using ```python```  format.

===task===
### Solution code
\end{Verbatim}
\end{tcolorbox}

\section{Examples}
\label{app:example}
\subsection{Generated Tools}

\textbf{The tools generated for the Open-ended Tasks are as follows:}
\begin{tcolorbox}[title=CraftDiamondHelmet, width=\textwidth,top=0mm,  breakable]
\begin{Verbatim}[breaklines=true, breakanywhere=true, fontsize=\footnotesize]
async function craftDiamondHelmet(bot) {
  const mcData = require('minecraft-data')(bot.version);
  const Vec3 = require('vec3').Vec3;

  // Check inventory for the number of diamonds
  const diamondCount = bot.inventory.count(mcData.itemsByName["diamond"].id);
  const requiredDiamonds = 5;

  // If not enough diamonds, mine diamond ores
  if (diamondCount < requiredDiamonds) {
    const diamondsToMine = requiredDiamonds - diamondCount;
    bot.chat(`Need ${diamondsToMine} more diamond(s). Mining...`);
    await mineBlock(bot, "diamond_ore", diamondsToMine);
  }

  // Ensure a crafting table is placed nearby
  let craftingTable = bot.findBlock({
    matching: mcData.blocksByName.crafting_table.id,
    maxDistance: 32
  });
  if (!craftingTable) {
    bot.chat("Placing crafting table...");
    await placeItem(bot, "crafting_table", bot.entity.position.offset(1, 0, 0));
    craftingTable = bot.findBlock({
      matching: mcData.blocksByName.crafting_table.id,
      maxDistance: 32
    });
  }

  // Craft the diamond helmet
  bot.chat("Crafting diamond helmet...");
  await craftItem(bot, "diamond_helmet", 1);
  bot.chat("Diamond helmet crafted successfully.");
}
\end{Verbatim}
\end{tcolorbox}

\begin{tcolorbox}[title=CraftItemWithMaterials, width=\textwidth,top=0mm,  breakable]
\begin{Verbatim}[breaklines=true, breakanywhere=true, fontsize=\footnotesize]
async function craftItemWithMaterials(bot, itemName, requiredMaterials) {
  const mcData = require('minecraft-data')(bot.version);
  const Vec3 = require('vec3').Vec3;

  // Check inventory for required materials
  for (const material of requiredMaterials) {
    let itemCount = bot.inventory.count(mcData.itemsByName[material.name].id);
    if (itemCount < material.count) {
      const requiredCount = material.count - itemCount;
      bot.chat(`Need ${requiredCount} more ${material.name}(s).`);
      if (material.name === "diamond") {
        let diamondOre = await bot.findBlock({
          matching: mcData.blocksByName["diamond_ore"].id,
          maxDistance: 32
        });
        if (!diamondOre) {
          bot.chat("No diamond ore found nearby. Exploring...");
          diamondOre = await exploreUntil(bot, new Vec3(1, 0, 1), 60, () => {
            return bot.findBlock({
              matching: mcData.blocksByName["diamond_ore"].id,
              maxDistance: 32
            });
          });
        }
        if (diamondOre) {
          await mineBlock(bot, "diamond_ore", requiredCount);
        } else {
          bot.chat("Failed to find diamond ore after exploring.");
          return;
        }
      } else if (material.name === "stick") {
        const woodenPlanksCount = bot.inventory.count(mcData.itemsByName["oak_planks"].id) + bot.inventory.count(mcData.itemsByName["birch_planks"].id);
        if (woodenPlanksCount < 2) {
          const requiredLogs = Math.ceil((2 - woodenPlanksCount) / 4);
          bot.chat(`Need more wooden planks. Gathering ${requiredLogs} logs...`);
          await obtainWoodLogs(bot, requiredLogs);
          await craftItem(bot, "oak_planks", requiredLogs);
        }
        bot.chat("Crafting sticks...");
        await craftItem(bot, "stick", 1);
      }
    }
  }

  // Ensure a crafting table is placed nearby
  let craftingTable = bot.findBlock({
    matching: mcData.blocksByName.crafting_table.id,
    maxDistance: 32
  });
  if (!craftingTable) {
    bot.chat("Placing crafting table...");
    await placeItem(bot, "crafting_table", bot.entity.position.offset(1, 0, 0));
    craftingTable = bot.findBlock({
      matching: mcData.blocksByName.crafting_table.id,
      maxDistance: 32
    });
  }

  // Craft the item
  bot.chat(`Crafting ${itemName}...`);
  await craftItem(bot, itemName, 1);
  bot.chat(`${itemName} crafted successfully.`);
}

async function craftDiamondAxe(bot) {
  const requiredMaterials = [{
    name: "diamond",
    count: 3
  }, {
    name: "stick",
    count: 2
  }];
  await craftItemWithMaterials(bot, "diamond_axe", requiredMaterials);
}
\end{Verbatim}
\end{tcolorbox}

\textbf{The tools generated for the Agent Tasks are as follows:}
Here, we can clearly see the call relationships between functions, thus forming more complex tools.
\begin{tcolorbox}[title=Tools for DA-Bench, width=\textwidth,top=0mm,  breakable]
\begin{Verbatim}[breaklines=true, breakanywhere=true, fontsize=\footnotesize]
def filter_rows_by_non_null(df: pd.DataFrame, column_name: str) -> pd.DataFrame:
    """
    Filters rows in a dataset based on non-null values in a specified column.
    
    Parameters:
    - df (pd.DataFrame): The input DataFrame.
    - column_name (str): The name of the column to filter by non-null values.
    
    Returns:
    - pd.DataFrame: A DataFrame with rows containing non-null values in the specified column.
    
    Raises:
    - ValueError: If the specified column is not found in the DataFrame.
    """
    # Check if the column exists in the DataFrame
    if column_name not in df.columns:
        raise ValueError(f"Column '{column_name}' not found in the DataFrame.")
    
    # Filter rows based on non-null values in the specified column
    filtered_df = df.dropna(subset=[column_name])
    
    return filtered_df

def convert_column_to_numeric(df: pd.DataFrame, column_name: str) -> pd.DataFrame:
    """
    Converts a specified column in a DataFrame to numeric values, handling non-numeric values appropriately.
    
    Parameters:
    - df (pd.DataFrame): The input DataFrame.
    - column_name (str): The name of the column to convert to numeric values.
    
    Returns:
    - pd.DataFrame: The DataFrame with the specified column converted to numeric values.
    
    Raises:
    - ValueError: If the specified column is not found in the DataFrame.
    """
    # Check if the column exists in the DataFrame
    if column_name not in df.columns:
        raise ValueError(f"Column '{column_name}' not found in the DataFrame.")
    
    # Convert the specified column to numeric values, setting non-numeric values to NaN
    df[column_name] = pd.to_numeric(df[column_name], errors='coerce')
    
    # Filter out rows with non-numeric values in the specified column using the existing tool
    df = filter_rows_by_non_null(df, column_name)
    
    return df

def create_sum_feature(df: pd.DataFrame, new_column_name: str, columns_to_sum: list) -> pd.DataFrame:
    """
    Creates a new feature by summing specified columns in a DataFrame.
    
    Parameters:
    - df (pd.DataFrame): The input DataFrame.
    - new_column_name (str): The name of the new column to be created.
    - columns_to_sum (list): A list of column names to sum.
    
    Returns:
    - pd.DataFrame: The DataFrame with the new feature added.
    
    Raises:
    - ValueError: If any of the specified columns are not found in the DataFrame.
    """
    # Check if all specified columns exist in the DataFrame
    for column in columns_to_sum:
        if column not in df.columns:
            raise ValueError(f"Column '{column}' not found in the DataFrame.")
    
    # Convert specified columns to numeric values
    for column in columns_to_sum:
        df = convert_column_to_numeric(df, column)
    
    # Create the new feature by summing the specified columns
    df[new_column_name] = df[columns_to_sum].sum(axis=1)
    
    return df
\end{Verbatim}
\end{tcolorbox}

\begin{tcolorbox}[title=Tools for TextCraft, width=\textwidth,top=0mm, breakable]
\begin{Verbatim}[breaklines=true, breakanywhere=true, fontsize=\footnotesize]
def gather_materials_for_dye(required_materials: dict) -> bool:
    """
    Gathers the required materials for crafting any dye.
    
    Parameters:
    - required_materials (dict): A dictionary where keys are material names and values are the required quantities.
    
    The tool checks the inventory for these materials and gathers them if they are missing.
    
    Returns:
    - bool: True if all materials were successfully gathered, False otherwise.
    """
    # Gather the required materials
    if not gather_materials(required_materials):
        return False
    
    # Check if we have white dye, if not gather bone meal or lily of the valley to craft it
    inventory = check_inventory()
    if "white dye" in required_materials and "white dye" not in inventory:
        if not gather_materials({"bone meal": 1}) and not gather_materials({"lily of the valley": 1}):
            return False
        # Craft white dye using bone meal or lily of the valley
        if "bone meal" in inventory:
            craft_object("1 white dye", ["1 bone meal"])
        elif "lily of the valley" in inventory:
            craft_object("1 white dye", ["1 lily of the valley"])
    
    # Recheck the inventory to ensure all materials are gathered
    missing_items = check_missing_items([f"{qty} {item}" for item, qty in required_materials.items()])
    if missing_items:
        print(f"Missing items: {missing_items}")
        return False
    
    # Successfully gathered all materials
    return True

def craft_orange_dye(quantity: int) -> bool:
    """
    Crafts the specified quantity of orange dye.
    
    Parameters:
    - quantity (int): The number of orange dye to craft.
    
    Returns:
    - bool: True if the orange dye was successfully crafted, False otherwise.
    """
    # Define the required materials for crafting orange dye
    required_materials = {"orange tulip": quantity, "red dye": quantity, "yellow dye": quantity}
    
    # Gather the required materials using the existing gather_materials_for_dye function
    if not gather_materials_for_dye(required_materials):
        return False
    
    # Check the inventory for available materials
    inventory = check_inventory()
    
    # Craft orange dye using orange tulip if available
    if "orange tulip" in inventory:
        craft_object(f"{quantity} orange dye", [f"{quantity} orange tulip"])
        print(f"Crafted {quantity} orange dye using {quantity} orange tulip")
        return True
    
    # Craft orange dye using red dye and yellow dye if available
    if "red dye" in inventory and "yellow dye" in inventory:
        craft_object(f"{quantity} orange dye", [f"{quantity} red dye", f"{quantity} yellow dye"])
        print(f"Crafted {quantity} orange dye using {quantity} red dye and {quantity} yellow dye")
        return True
    
    # If neither method was successful, return False
    print("Failed to craft orange dye.")
    return False
\end{Verbatim}
\end{tcolorbox}

\textbf{The tools generated for the Single-turn Code Task are as follows:}
\begin{tcolorbox}[title=Tools for MATH, width=\textwidth,top=0mm, breakable]
\begin{Verbatim}[breaklines=true, breakanywhere=true, fontsize=\footnotesize]
def find_integer_satisfying_condition(condition):
    """
    Find the smallest positive integer that satisfies the given condition.

    Parameters:
        condition (function): A lambda function representing the condition to be checked.

    Returns:
        int: The smallest positive integer that satisfies the condition.
    """
    x = 1
    while True:
        if condition(x):
            return x
        x += 1

def calculate_min_correct_answers(total_problems, passing_percentage):
    """
    Calculate the minimum number of correct answers required to pass a test based on the total number of problems and the passing percentage.

    Parameters:
        total_problems (int): The total number of problems on the test.
        passing_percentage (float): The passing percentage required to pass the test.

    Returns:
        int: The minimum number of correct answers required to pass the test.
    """
    if total_problems <= 0:
        return "Total number of problems must be greater than zero."
    if not (0 <= passing_percentage <= 100):
        return "Passing percentage must be between 0 and 100."

    required_correct_answers = (passing_percentage / 100) * total_problems

    # Use find_integer_satisfying_condition to find the minimum integer satisfying the condition
    min_correct_answers = find_integer_satisfying_condition(lambda x: x >= required_correct_answers)
    
    return min_correct_answers
\end{Verbatim}
\end{tcolorbox}

\begin{tcolorbox}[title=Tools for Date, width=\textwidth,top=0mm, breakable]
\begin{Verbatim}[breaklines=true, breakanywhere=true, fontsize=\footnotesize]
def calculate_date_by_days(start_date_str: str, days_to_add: int, date_format="%m/%d/%Y") -> str:
    """
    Calculates the date a specified number of days before or after a given date.

    Parameters:
    - start_date_str (str): The starting date as a string in the format MM/DD/YYYY.
    - days_to_add (int): The number of days to add (positive) or subtract (negative) from the start date.
    - date_format (str): The format of the input and output date string. Default is 'MM/DD/YYYY'.

    Returns:
    - str: The resulting date in the format MM/DD/YYYY.
    
    Raises:
    - ValueError: If the input date string does not match the specified format.
    - OverflowError: If the resulting date is out of the valid range for dates.
    """
    from datetime import datetime, timedelta

    try:
        # Parse the input date string into a date object using the provided format
        start_date = datetime.strptime(start_date_str, date_format).date()

        # Calculate the new date by adding the specified number of days
        new_date = start_date + timedelta(days=days_to_add)

        # Format the new date back into the desired string format
        result_date_str = new_date.strftime(date_format)

        return result_date_str
    except ValueError as e:
        raise ValueError("Incorrect date format. Please ensure the date string matches the provided format.") from e
    except OverflowError as e:
        raise OverflowError("The resulting date is out of the valid range for dates.") from e

def calculate_date_by_days_uk_format(start_date_str: str, days_to_add: int) -> str:
    """
    Calculates the date a specified number of days before or after a given date in UK format (DD/MM/YYYY).

    Parameters:
    - start_date_str (str): The starting date as a string in the format DD/MM/YYYY.
    - days_to_add (int): The number of days to add (positive) or subtract (negative) from the start date.

    Returns:
    - str: The resulting date in the format MM/DD/YYYY.
    
    Raises:
    - ValueError: If the input date string does not match the specified format.
    """
    from datetime import datetime

    try:
        # Convert the input date from DD/MM/YYYY to MM/DD/YYYY
        start_date = datetime.strptime(start_date_str, "%d/%m/%Y")
        
        # Use the existing tool to calculate the new date
        result_date_str = calculate_date_by_days(start_date.strftime("%m/%d/%Y"), days_to_add, "%m/%d/%Y")
        
        return result_date_str
    except ValueError as e:
        raise ValueError("Incorrect date format. Please ensure the date string matches the provided format.") from e
\end{Verbatim}
\end{tcolorbox}

\begin{tcolorbox}[title=Tools for TabMWP, width=\textwidth,top=0mm, breakable]
\begin{Verbatim}[breaklines=true, breakanywhere=true, fontsize=\footnotesize]
import pandas as pd

def stem_and_leaf_to_dataframe(stem_leaf_dict: dict) -> pd.DataFrame:
    """
    Converts a stem-and-leaf plot into a DataFrame.

    Parameters:
    - stem_leaf_dict (dict): A dictionary where keys are the stems and values are lists of leaves.

    Returns:
    - pd.DataFrame: A DataFrame with a single column containing the combined values of stems and leaves.
    """
    # Initialize an empty list to store the combined values
    combined_values = []

    # Iterate through the dictionary to combine stems and leaves
    for stem, leaves in stem_leaf_dict.items():
        for leaf in leaves:
            combined_value = int(f"{stem}{leaf}")
            combined_values.append(combined_value)

    # Create a DataFrame from the combined values
    df = pd.DataFrame(combined_values, columns=["Values"])
    
    return df

import pandas as pd

def count_value_occurrences(stem_leaf_dict: dict, value) -> int:
    """
    Counts the occurrences of a specific value in a DataFrame column created from a stem-and-leaf plot.

    Parameters:
    - stem_leaf_dict (dict): A dictionary where keys are the stems and values are lists of leaves.
    - value: The value to count in the DataFrame.

    Returns:
    - int: The count of the specified value in the DataFrame.
    """
    # Convert the stem-and-leaf plot to a DataFrame using the existing tool
    df = stem_and_leaf_to_dataframe(stem_leaf_dict)
    
    # Count the occurrences of the specified value in the DataFrame
    count = df["Values"].value_counts().get(value, 0)
    
    return count
\end{Verbatim}
\end{tcolorbox}

\end{document}